\documentclass{article} 
\usepackage{iclr2027_conference,times}

\usepackage{amsmath,amsfonts,bm}

\def\eqref#1{equation~\ref{#1}}

\def\1{\bm{1}}

\DeclareMathAlphabet{\mathsfit}{\encodingdefault}{\sfdefault}{m}{sl}
\SetMathAlphabet{\mathsfit}{bold}{\encodingdefault}{\sfdefault}{bx}{n}

\usepackage{hyperref}
\usepackage{url}
\usepackage{amsmath}
\usepackage{graphicx}
\usepackage[table]{xcolor}
\usepackage{placeins}
\usepackage{hyperref}
\usepackage{wrapfig}
\usepackage{siunitx}
\usepackage{float}
\usepackage[utf8]{inputenc} 
\usepackage[T1]{fontenc}    
\usepackage{hyperref}       
\usepackage[capitalise]{cleveref}
\usepackage{url}            
\usepackage{booktabs}       
\usepackage{amsfonts}       
\usepackage{nicefrac}       
\usepackage{microtype}      
\usepackage{xcolor}         
\usepackage{multirow}
\usepackage{natbib}
\usepackage{enumitem}
\usepackage{soul}
\usepackage{algorithm}
\usepackage{algpseudocode}
\usepackage{xcolor}
\definecolor{mydarkblue}{rgb}{0,0.08,0.45}
\hypersetup{
  colorlinks=true,
  linkcolor=mydarkblue,
  citecolor=mydarkblue,
  urlcolor=mydarkblue
}
\usepackage{wrapfig}
\algrenewcommand{\algorithmiccomment}[1]{\hfill$\triangleright$\ \textit{\textcolor{gray!70}{#1}}}

\title{Advancing the Calibration-Compute Frontier \\of LLM Classifiers via Asymmetric Duos}

\author{Berke Akturk$^*$, Ali Can Sahin$^*$, Rasit Cem Solatay$^*$ \\
Department of Statistics, LMU Munich \\
Munich, Germany
\AND 
Emanuel Sommer, Yawei Li, David Rügamer\\
Department of Statistics, LMU Munich \\
Munich Center for Machine Learning (MCML) \\
Munich, Germany
}

\iclrfinalcopy 
\begin{document}

\maketitle
\begingroup
  \renewcommand\thefootnote{*}
  \footnotetext{Equal contribution.}
\endgroup

\begin{abstract}
  The importance of reliable uncertainty estimation in large language models (LLMs) has become increasingly pronounced, yet modern models remain poorly calibrated and frequently overconfident. In this study, we address this by proposing an efficient method for LLM classifier calibration using the \emph{Asymmetric Duos} framework. Our approach combines a strong base model with a smaller sidekick model via logit-space fusion, utilizing global scaling parameters optimized on a validation set. Across diverse benchmarks, this Asymmetric LLM Duo consistently achieves best or second-best performance on expected calibration error and negative log-likelihood, without degrading predictive accuracy. More importantly, these gains are realized at a fraction of the computational cost of more complex baselines, placing our method on the calibration-compute Pareto frontier. We further demonstrate that our method acts orthogonal to various established calibration-aware techniques, meaning it can be easily combined with them to consistently boost their performance. Finally, our analysis of the logit-space behavior illustrates how asymmetric fusion successfully mitigates base model overconfidence. Overall, our findings demonstrate that the benefits of an asymmetric fusion successfully transfer to the language domain, providing a simple, highly cost-efficient approach to improving uncertainty estimation in LLM classifiers.

\end{abstract}

\section{Introduction}

Large language models (LLMs) have demonstrated remarkable predictive performance across a wide range of tasks and domains~\citep{brown2020gpt3, openai2024gpt4, llama2, llama3, minaee2024llm_survey}. These capabilities can be further increased for downstream tasks through fine-tuning methods~\citep{hu2022lora, yang2024bayesian, wang2024blob, lialin2024peftsurvey, wang2022list}. Despite the strong predictive performance of LLMs, these models often remain poorly calibrated and exhibit overconfidence~\citep{guo2017calibration, xiong2024confidence, leng2025overconfidence, shorinwa2024uqsurvey}, and fine-tuning procedures can further amplify this behavior~\citep{jiang2020calibration_lm}. LLMs being overconfident in their predictions can yield serious problems for critical decision-making domains~\citep{kumar2023conformal_llm}, and being overconfident can also undermine the reliability and restrict the use cases of LLMs in those domains~\citep{li2025calibrating_llm}. Therefore, in real-world applications, these neural networks should also be able to indicate when they can be incorrect~\citep{guo2017calibration}. This capability is known as uncertainty quantification or confidence calibration.

\begin{figure}[htbp]
    \centering
    \includegraphics[width=1\linewidth]{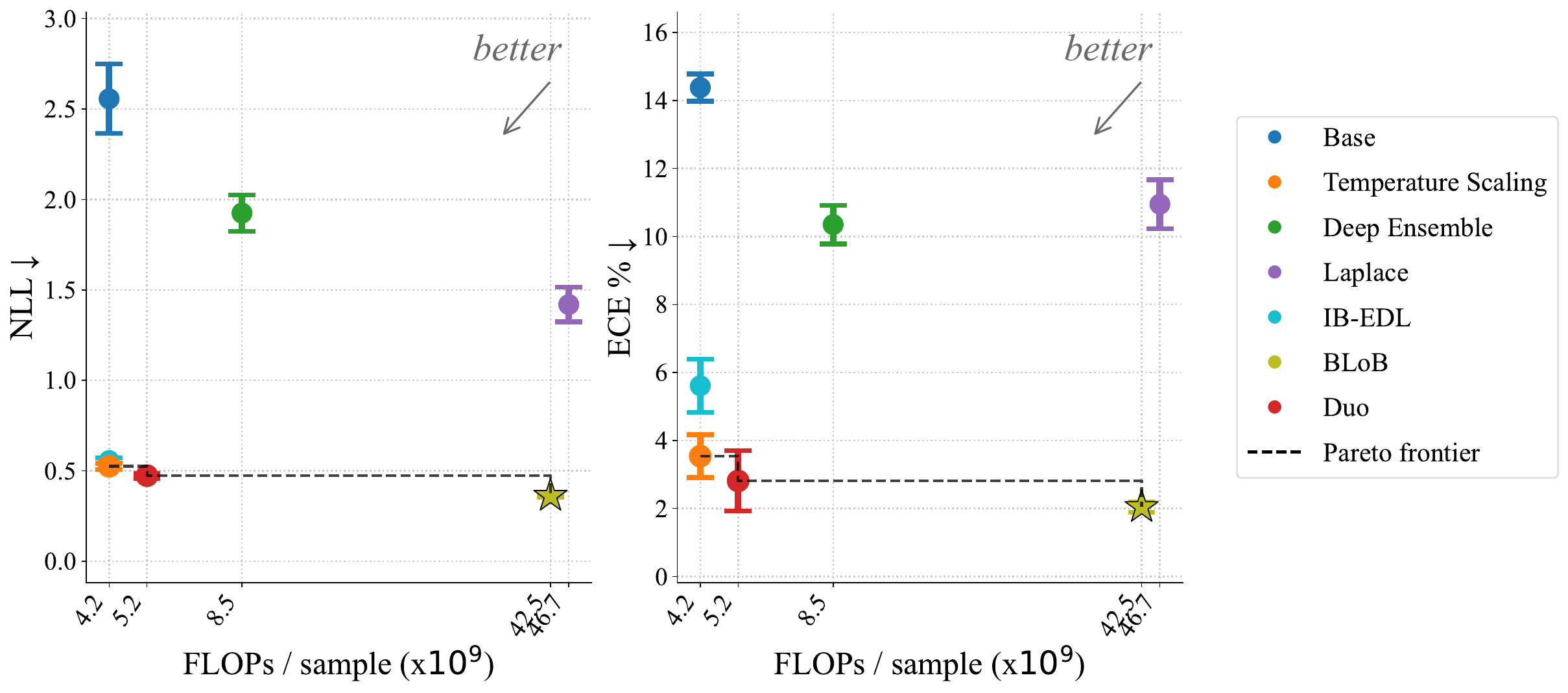}
    \caption{Pareto frontier of floating-point operations (FLOPs) vs.\ negative log-likelihood (NLL; left) and expected calibration error (ECE; right) for the CSQA dataset. For both metrics, Temperature Scaling, Duo, and BLoB lie on Pareto frontier. The method with the best NLL/ECE performance is marked with a star.}
    \label{fig:pareto_csqa}
\end{figure}

Overconfidence is not specific to LLMs. Modern neural networks are often poorly calibrated and exhibit overconfidence~\citep{guo2017calibration}. 
One solution to this is the recently proposed \emph{Asymmetric Duo} \citep{zhou2025asymmetric_duos}. They showed that using a smaller sidekick neural network model together with a larger base model can improve the uncertainty estimates of those neural networks. Their results suggest that uncertainty estimation can be improved with only a modest computational overhead relative to the larger base model.

Since \citet{zhou2025asymmetric_duos} exclusively focused on computer vision tasks and models, it remains unclear whether and how Asymmetric Duos can work for LLM classifiers. In this paper, we pursue this question for LLMs fine-tuned to multiple-choice question answering (MCQA). Specifically, we examine whether asymmetric fusion improves the calibration of LLMs in MCQA and whether it changes the predictive behavior of the base model in a systematic and cost-efficient way. This positions the paper at the intersection of LLM calibration, uncertainty-aware adaptation, and efficient multi-model uncertainty estimation. Our contributions are organized as follows:

\begin{enumerate}[leftmargin=0.8cm]
    \item \textbf{Cost-efficient calibration:} We show that Asymmetric Duos improve LLM calibration on the cost--performance frontier, matching or surpassing state-of-the-art baselines often at a fraction of the inference FLOPs across all benchmarks (cf. Figure~\ref{fig:pareto_csqa}).
    \item \textbf{Compatibility with calibration-aware methods:} We show that Asymmetric Duos compose additively with existing calibration-aware approaches such as BLoB, Laplace-LoRA, and IB-EDL, consistently improving calibration performance when applied in composition with these methods.
    \item \textbf{Logit-space mechanism:} Our mechanistic study sheds light on how asymmetric logit fusion reshapes the logit geometry of the base model, reducing overconfidence by smoothing extreme predictions while preserving decision boundaries, thereby improving calibration without degrading accuracy.
\end{enumerate} 

\section{Related Work \& Background}

\paragraph{Calibration and uncertainty in neural networks and LLMs.}

The objective of confidence calibration is to match the confidence assigned by a model with the empirical probability of correctness. Although modern neural networks achieve strong predictive performance, this does not necessarily imply well-calibrated confidence estimates, and overconfidence remains a well-known issue~\citep{guo2017calibration, jiang2020calibration_lm, shorinwa2024uqsurvey}. This has motivated substantial research on uncertainty estimation and post-hoc calibration, with temperature scaling being one of the most widely used and computationally simple approaches~\citep{guo2017calibration}. Evidential methods instead aim to capture predictive uncertainty directly within the model~\citep{sensoy2018evidential_dl}.

Despite their strong performance, LLMs may produce unreliable uncertainty estimates, particularly after adaptation to downstream tasks~\citep{he2023preserving, li2025calibrating_llm}. Calibration is also affected by prompting, decoding strategies, and answer extraction, making it more complex than in classical neural networks~\citep{zhang2024elicitingfidelity}.

\paragraph{Multi-model uncertainty estimation and Asymmetric Duos.}

Ensemble-based approaches improve uncertainty estimation by combining predictions from multiple models, but standard deep ensembles are computationally expensive because they require training and evaluating several models for each prediction~\citep{lakshminarayanan2017deep_ensembles, fort2019deepens}. This cost has motivated more efficient uncertainty-aware alternatives. In the LLM setting, recent work has focused on uncertainty estimation through LoRA ensembles and Bayesian parameter-efficient adaptation methods. For example, Laplace-LoRA applies a Laplace approximation to LoRA adapters for post-hoc Bayesian inference~\citep{yang2024bayesian, daxberger2021laplace}, while BLoB performs Bayesian LoRA through variational inference during fine-tuning~\citep{wang2024blob}. Ensemble-based approaches for LoRA-adapted LLMs have also been explored~\citep{balabanov2025loraensembles, wang2023loraensembles}.

Further background on MCQA as a testbed for uncertainty estimation and on the relationship between parameter-efficient fine-tuning and calibration-aware training is provided in Appendix~\ref{app:additional_background}.


\section{Asymmetric LLM Duos}
\label{sec:duos}

Our work builds on the idea of \citet{zhou2025asymmetric_duos} by studying whether asymmetric logit fusion transfers to LLM-based MCQA. In addition, we investigate whether the Asymmetric Duos setup can further improve performance when combined with existing calibration-aware methods such as BLoB and Laplace-LoRA.

\subsection{Problem formulation}
\label{sec:duo_problem}

Let \(\mathbf{x} \in \mathcal{X}\) denote the input and \(y \in \mathcal{Y}\) the corresponding label, where \(|\mathcal{Y}| = K\). We consider two predictive models: a base model \(f_{\text{base}}\) and a sidekick model \(f_{\text{side}}\). Each model produces a logit vector over the label space, \(\mathbf{z}_{\text{base}}(\mathbf{x}) \in \mathbb{R}^K\) and \(\mathbf{z}_{\text{side}}(\mathbf{x}) \in \mathbb{R}^K\), which yield class probability vectors \(p_{\text{base}}(y \mid \mathbf{x}) = \mathrm{softmax}(\mathbf{z}_{\text{base}}(\mathbf{x}))\) and \(p_{\text{side}}(y \mid \mathbf{x}) = \mathrm{softmax}(\mathbf{z}_{\text{side}}(\mathbf{x}))\). Our goal is to construct a combined predictor that improves the quality of uncertainty estimates while maintaining predictive performance.

Following~\citet{zhou2025asymmetric_duos}, we define the Asymmetric Duo predictor via a linear combination in logit space, $\mathbf{z}_{\text{duo}}(\mathbf{x}) = \alpha\,\mathbf{z}_{\text{base}}(\mathbf{x}) + \beta\,\mathbf{z}_{\text{side}}(\mathbf{x})$. These combined logits parameterize a categorical distribution over answer choices, with class probabilities given by $p_{\text{duo}}(y \mid \mathbf{x}) = \mathrm{softmax}(\mathbf{z}_{\text{duo}}(\mathbf{x}))$. The final prediction is obtained as $\hat{y}(\mathbf{x})=\arg\max_{y\in\mathcal{Y}}p_{\mathrm{duo}}(y\mid\mathbf{x})$. The weights \(\alpha\) and \(\beta\) are global parameters shared across all inputs. We deliberately restrict the fusion to global, input-independent weights in order to avoid introducing additional model complexity and training overhead.
\begin{algorithm}[!t]
\caption{Asymmetric Duo training and inference}
\label{alg:duo}
\begin{algorithmic}[1]
\Require $\mathcal{D}_{\text{train}}, \mathcal{D}_{\text{val}}, \mathcal{D}_{\text{test}}$, 
models $f_{\text{base}}, f_{\text{side}}$

\State Train $f_{\text{base}}, f_{\text{side}}$ on $\mathcal{D}_{\text{train}}$ 
\Comment{Independent fine-tuning}

\For{each $(\mathbf{x}_i, y_i) \in \mathcal{D}_{\text{val}}$}  
    \State $\mathbf{z}_{\text{base}}^{(i)} \gets f_{\text{base}}(\mathbf{x}_i)$
    \State $\mathbf{z}_{\text{side}}^{(i)} \gets f_{\text{side}}(\mathbf{x}_i)$
    \Comment{Logit extraction}
\EndFor

\State
$(\alpha^*, \beta^*) = \arg\min_{\alpha, \beta \ge 0} \mathcal{L}(\alpha, \beta)$ 
\Comment{Weight optimization (\cref{eq:duo_nll,eq:duo_argmin})}

\For{each $\mathbf{x} \in \mathcal{D}_{\text{test}}$} \Comment{Inference}    
    \State $\mathbf{z}_{\text{duo}}(\mathbf{x}) \gets \alpha^* \mathbf{z}_{\text{base}}(\mathbf{x}) + \beta^* \mathbf{z}_{\text{side}}(\mathbf{x})$
    \State $\hat{y} \gets \arg\max_{y \in \mathcal{Y}} \ \mathrm{softmax}(\mathbf{z}_{\text{duo}}(\mathbf{x}))$
\EndFor

\end{algorithmic}
\end{algorithm}
\subsection{Learning the fusion weights}
\label{sec:duo_learning}

Let \(\mathcal{D}_{\text{val}} = \{(\mathbf{x}_i, y_i)\}_{i=1}^N\) denote the validation set. For each validation example, we precompute and store the aligned logit vectors \(\mathbf{z}_{\text{base}}^{(i)}\) and \(\mathbf{z}_{\text{side}}^{(i)}\). We then estimate the fusion scales \(\alpha\) and \(\beta\) by solving the optimization problem
\begin{equation}
(\alpha^*, \beta^*)
=
\arg\min_{\alpha, \beta \ge 0}
\mathcal{L}(\alpha, \beta),
\label{eq:duo_argmin}
\end{equation}
where the validation negative log-likelihood (NLL) is defined as
\begin{equation}
\mathcal{L}(\alpha, \beta)
=
-\frac{1}{N}\sum_{i=1}^{N}
\log
\left[
\mathrm{softmax}
\left(
\alpha \mathbf{z}_{\text{base}}^{(i)}
+
\beta \mathbf{z}_{\text{side}}^{(i)}
\right)
\right]_{y_i}.
\label{eq:duo_nll}
\end{equation}

In our implementation, this is a two-dimensional post-hoc optimization problem solved with L-BFGS-B, a quasi-Newton method for bound-constrained optimization~\citep{zhu1997lbfgsb}. The fusion scales are not constrained to sum to one. Instead, we only enforce non-negativity through lower bounds on \(\alpha\) and \(\beta\). Thus, \(\alpha\) and \(\beta\) should be interpreted as effective logit scales rather than normalized mixture weights. 

The logits of both the base and sidekick models are matched by example index before optimization, ensuring that the two models are combined on the same inputs. Since this procedure uses only saved validation predictions, the Duo can be constructed after independent fine-tuning without any joint training of the two models; \Cref{alg:duo} gives the complete training and inference pipeline.

\subsection{Interpretation}
\label{sec:duo_interpretation}
The learned weights $\alpha$ and $\beta$ determine the relative contributions of the base and sidekick models across the dataset. Since the optimization is driven by validation NLL, the resulting combination reflects which model provides more informative or better-calibrated logits. If the sidekick model provides little useful information, the optimization assigns it a small weight, effectively recovering the base model. Conversely, when the sidekick model captures complementary patterns, its contribution increases and improves the Duo predictor. In our experiments, we empirically validate this behavior in Table~\ref{tab:results_overall}, where we observe consistent improvements in NLL and expected calibration error (ECE) across all datasets. 

\paragraph{Implicit per-model temperature scaling.}
Beyond inter-model weighting, the unconstrained non-negative parameterization of $(\alpha,\beta)$ has a direct calibration interpretation: each scale acts as an inverse temperature on its own model's logits. Concretely, $\alpha\,\mathbf{z}_{\text{base}}(\mathbf{x}) = \mathbf{z}_{\text{base}}(\mathbf{x})/T_{\text{base}}$ with $T_{\text{base}} = 1/\alpha$, and analogously for the sidekick model. The validation-NLL minimization in \cref{eq:duo_argmin} therefore performs two operations in a single two-dimensional fit: (i) per-model temperature scaling~\citep{guo2017calibration}, and (ii) inter-model relative weighting. This is in contrast with formulations that fix $\alpha + \beta = 1$, which pin the joint logit magnitude and decouple it from any temperature effect. By keeping $(\alpha,\beta)$ unconstrained on the non-negative orthant, Duo includes both calibration steps without an extra optimization stage. Consequently, a Duo built from temperature-scaled base and sidekick logits, $\mathbf{z}_{\text{duo}}(\mathbf{x}) = \alpha\,\mathbf{z}_\text{base}(\mathbf{x})/T_\text{base} + \beta\,\mathbf{z}_\text{side}(\mathbf{x})/T_\text{side}$, can always be reparameterized as $\tilde{\alpha}\,\mathbf{z}_\text{base}(\mathbf{x}) + \tilde{\beta}\,\mathbf{z}_\text{side}(\mathbf{x})$ with $\tilde{\alpha}=\alpha/T_\text{base}$ and $\tilde{\beta}=\beta/T_\text{side}$, effectively recovering the same predictor as the plain Duo.

\paragraph{Gains beyond temperature scaling.} To separate what a single temperature achieves from what the Sidekick adds, we evaluate each base method (plain LoRA, LA, and IB-EDL) in three forms: its original logits (Base), Base TS with one learned scale, and Duo with two learned scales. Base TS $-$ Base then measures temperature scaling alone, and Duo $-$ Base TS the additional contribution of the sidekick (Table~\ref{tab:duo_vs_temperature_scaling}). For plain LoRA and LA, temperature scaling provides most of the NLL improvement (e.g., $-1.579$ on ARC-C for plain LoRA), and Duo adds a further $0.003$--$0.054$. For IB-EDL, whose evidential objective already calibrates the temperature, TS changes NLL by at most $0.014$, yet Duo still reduces it by up to $0.064$ (CSQA). Since Duo improves on Base TS in all $18$ method--dataset combinations, the Sidekick contributes information that no rescaling of Base can provide.
\section{Experiments}
\label{sec:exp}

We now study our adaptation of the Asymmetric Duo for LLM MCQA.

\paragraph{Setup.} Unless stated otherwise, we use Qwen2-7B as Base (B) and Qwen2-1.5B as Sidekick (S)~\citep{qwen2}, both fine-tuned with LoRA~\citep{hu2022lora}, on six MCQA benchmarks: ARC-C and ARC-E~\citep{arc}, OpenBookQA (OBQA)~\citep{obqa}, CommonsenseQA (CSQA)~\citep{csqa}, SciQ~\citep{sciq}, and RACE~\citep{race}. We compare against the fine-tuned Base (B), its temperature-scaled version (B TS)~\citep{guo2017calibration}, a two-member deep ensemble (B DE)~\citep{lakshminarayanan2017deep_ensembles}, Laplace-LoRA (B LA)~\citep{yang2024bayesian}, IB-EDL (B IB-EDL)~\citep{li2025calibrating_llm}, and BLoB (B BLoB)~\citep{wang2024blob}. An effective uncertainty-aware method should both enhance model calibration and maintain predictive accuracy comparable to standard MAP training~\citep{li2025calibrating_llm}. Therefore, we use accuracy (Acc), ECE, and NLL as mean $\pm$ SD over five seeds, and measure inference cost in FLOPs, summing Base and Sidekick for Duo. Further details are given in Appendix~\ref{app:det_imp}.

\subsection{Cost-efficient uncertainty quantification}
\label{sec:pareto}

\begin{table}[!t]
\caption{Calibration performance of baseline and Duo methods fine-tuned on Qwen2 7B (``B'' for Base) and Qwen2 1.5B (``S'' for Sidekick). Arrows ($\uparrow$ or $\downarrow$) indicate the direction of better performance for each metric. The best results are highlighted in \textbf{bold}, and the second-best results are \underline{underlined}. Additionally, $\pm$ denotes the standard deviation over 5 runs with 5 different seeds. The Duo method has either the best or the second-best performance across most of the datasets and metrics.}

\centering
\footnotesize
\setlength{\tabcolsep}{5pt}
\label{tab:results_overall}
\resizebox{\textwidth}{!}{
\begin{tabular}{llcccccc}
\toprule
Metric & Method & ARC-C & ARC-E & CSQA & OBQA & RACE & SciQ \\
\midrule
\multirow{8}{*}{NLL $\downarrow$}
  & S                                & $4.234_{\pm0.046}$ & $1.968_{\pm0.100}$ & $3.493_{\pm0.182}$ & $1.497_{\pm0.076}$ & $0.590_{\pm0.021}$ & $0.546_{\pm0.022}$ \\
  & B                                & $2.075_{\pm0.041}$ & $0.655_{\pm0.065}$ & $2.557_{\pm0.192}$ & $0.990_{\pm0.080}$ & $0.340_{\pm0.009}$ & $0.333_{\pm0.018}$ \\
  & B TS                             & \underline{$0.496_{\pm0.011}$} & $0.213_{\pm0.015}$ & $0.525_{\pm0.017}$ & $0.357_{\pm0.012}$ & $0.268_{\pm0.005}$ & $0.183_{\pm0.008}$ \\
  & B DE                          & $1.682_{\pm0.069}$ & $0.538_{\pm0.029}$ & $1.925_{\pm0.100}$ & $0.750_{\pm0.054}$ & $0.319_{\pm0.006}$ & $0.295_{\pm0.007}$ \\
  & B LA                             & $1.212_{\pm0.052}$ & $0.401_{\pm0.029}$                 & $1.419_{\pm0.096}$ & $0.602_{\pm0.068}$ & $0.308_{\pm0.006}$                 & $0.259_{\pm0.016}$                 \\
  & B IB-EDL                         & $0.547_{\pm0.018}$ & $0.220_{\pm0.021}$ & $0.556_{\pm0.017}$ & $0.368_{\pm0.016}$ & $0.267_{\pm0.003}$ & $0.205_{\pm0.005}$ \\
  & B BLoB                           & $1.347_{\pm0.101}$ & $\mathbf{0.183_{\pm0.039}}$ & $\mathbf{0.361_{\pm0.008}}$ & $\mathbf{0.240_{\pm0.016}}$ & $\mathbf{0.247_{\pm0.009}}$ & $\mathbf{0.136_{\pm0.002}}$ \\
  & \cellcolor{gray!15}Duo                              &\cellcolor{gray!15} $\mathbf{0.469_{\pm0.009}}$ &\cellcolor{gray!15} \underline{$0.188_{\pm0.014}$}
  &\cellcolor{gray!15} \underline{$0.473_{\pm0.014}$} &\cellcolor{gray!15}  \underline{$0.302_{\pm0.007}$} &\cellcolor{gray!15} \underline{$0.262_{\pm0.006}$} &\cellcolor{gray!15} \underline{$0.167_{\pm0.006}$} \\
  \midrule
\multirow{8}{*}{Acc $\uparrow$}
  & S                                & $71.54_{\pm0.90}$ & $85.51_{\pm0.61}$ & $74.63_{\pm0.27}$ & $80.32_{\pm0.52}$ & $84.29_{\pm0.42}$ & $90.04_{\pm0.15}$ \\
  & B                                & $86.59_{\pm0.52}$ & $95.08_{\pm0.40}$ & $85.34_{\pm0.41}$ & $90.24_{\pm0.17}$ & $\underline{91.89_{\pm0.11}}$ & $94.96_{\pm0.21}$ \\
  & B TS                             & $86.59_{\pm0.52}$ & $95.08_{\pm0.40}$ & $85.34_{\pm0.41}$ & $90.24_{\pm0.17}$ & $\underline{91.89_{\pm0.11}}$ & $94.96_{\pm0.21}$ \\
  & B DE                          & $\mathbf{87.15_{\pm0.43}}$ & $95.09_{\pm0.14}$ & $\underline{85.81_{\pm0.35}}$ & $90.20_{\pm0.31}$ & $\mathbf{91.94_{\pm0.23}}$ & $95.12_{\pm0.25}$ \\
  & B LA                             & $86.66_{\pm0.48}$ & $95.11_{\pm0.32}$ & $85.32_{\pm0.42}$ & $90.04_{\pm0.14}$ & $91.80_{\pm0.15}$ & $94.84_{\pm0.23}$ \\
  & B IB-EDL                         & $86.28_{\pm0.60}$ & $\mathbf{95.23_{\pm0.17}}$ & $85.39_{\pm0.52}$ & $\underline{90.84_{\pm0.38}}$ & $91.85_{\pm0.26}$ & $95.02_{\pm0.22}$ \\
  & B BLoB                           & $86.31_{\pm0.67}$ & $94.89_{\pm0.38}$ & $\mathbf{86.55_{\pm0.61}}$ & $\mathbf{91.84_{\pm0.83}}$ & $91.57_{\pm0.59}$ & $\mathbf{95.42_{\pm0.08}}$ \\
  & \cellcolor{gray!15}Duo
  & \cellcolor{gray!15}$\underline{86.81_{\pm0.49}}$
  & \cellcolor{gray!15}$\underline{95.14_{\pm0.40}}$
  & \cellcolor{gray!15}$85.73_{\pm0.42}$
  & \cellcolor{gray!15}$90.60_{\pm0.14}$
  & \cellcolor{gray!15}$91.82_{\pm0.14}$
  & \cellcolor{gray!15}$\underline{95.24_{\pm0.26}}$ \\
  \midrule
\multirow{8}{*}{ECE $\downarrow$}
  & S                                & $27.73_{\pm0.96}$ & $14.16_{\pm0.59}$ & $24.61_{\pm0.36}$ & $18.57_{\pm0.65}$ & $10.11_{\pm0.47}$ & $8.69_{\pm0.14}$  \\
  & B                                & $13.05_{\pm0.52}$ & $4.74_{\pm0.42}$  & $14.38_{\pm0.40}$ & $9.55_{\pm0.32}$  & $5.63_{\pm0.15}$  & $4.75_{\pm0.13}$  \\
  & B TS                             & $\underline{4.35_{\pm0.84}}$  & $\underline{1.53_{\pm0.21}}$  & $3.54_{\pm0.63}$  & $4.13_{\pm1.07}$  & $\underline{2.73_{\pm0.30}}$  & $3.07_{\pm0.53}$  \\
  & B DE                          & $10.11_{\pm0.93}$ & $3.76_{\pm0.13}$  & $10.35_{\pm0.57}$ & $7.14_{\pm0.22}$  & $5.17_{\pm0.25}$  & $4.12_{\pm0.07}$  \\
  & B LA                             & $9.79_{\pm0.39}$ & $3.55_{\pm0.31}$ & $10.95_{\pm0.72}$ & $7.03_{\pm0.42}$ & $4.77_{\pm0.23}$ & $3.81_{\pm0.25}$ \\
  & B IB-EDL                         & $4.88_{\pm1.04}$  & $3.90_{\pm1.01}$  & $5.61_{\pm0.78}$  & $9.37_{\pm1.20}$  & $3.84_{\pm0.55}$  & $2.89_{\pm0.65}$  \\
  & B BLoB                           & $12.27_{\pm0.67}$ & $5.48_{\pm2.90}$  & $\mathbf{2.04_{\pm0.15}}$  & $\mathbf{2.71_{\pm0.75}}$  & $\mathbf{2.17_{\pm0.31}}$  & $\mathbf{1.55_{\pm0.38}}$  \\
  & \cellcolor{gray!15}Duo
  & \cellcolor{gray!15}$\mathbf{3.90_{\pm1.44}}$
  & \cellcolor{gray!15}$\mathbf{1.30_{\pm0.15}}$
  & \cellcolor{gray!15}$\underline{2.81_{\pm0.89}}$
  & \cellcolor{gray!15}$\underline{3.09_{\pm0.53}}$
  & \cellcolor{gray!15}$2.78_{\pm0.37}$
  & \cellcolor{gray!15}$\underline{1.94_{\pm0.42}}$ \\
\bottomrule
\end{tabular}}
\end{table}

As shown in Table~\ref{tab:results_overall}, Duo is highly competitive across all metrics and datasets compared to baselines. Duo achieves the best NLL on the ARC-C dataset, providing an almost $78\%$ decrease on average compared to Base ($2.075 \to 0.469$). For the other five datasets, Duo achieves the second best performance, with only small gaps relative to the best performing method; for example, on ARC-E, B BLoB obtains an average NLL of $0.183$, while Duo achieves $0.188$ and their confidence intervals overlap. Calibration gains also hold for ECE. Duo provides a reduction of approximately $60-80\%$ compared to Base. Additionally, for five of the six datasets, Duo is either the best or second-best-performing method, with RACE being the only dataset on which it does not achieve this ranking. However, its standard deviation band overlaps with that of the second-best-performing method, which is B TS. Despite consistently improving the calibration, Duo also preserves the predictive accuracy of Base. Across all datasets, Duo either improves accuracy or maintains it at a comparable level. From these perspectives, it can be considered as an effective method of predicting uncertainty.

\paragraph{Other model pairings.} To test whether these results depend on the Qwen2-7B/1.5B pairing, we repeat the experiment with three further pairings, comparing against B, B TS, and B DE (Tables~\ref{tab:additional_experiments_smaller_sidekick}--\ref{tab:additional_experiments_cross_family}); LA, IB-EDL, and BLoB are omitted because each requires method-specific training or posterior fitting for every model and dataset. (i)~With a smaller Sidekick, Qwen2-0.5B, Duo still attains the lowest NLL and ECE on all six datasets; its NLL gain over B TS is about $30$--$50\%$ of that obtained with the 1.5B Sidekick, so a larger Sidekick helps but is not required. (ii)~With a second model family, Llama-3.1-8B Base and Llama-3.2-1B Sidekick, Duo attains the lowest NLL on all six datasets and the lowest ECE on five (within one SD of B TS on ARC-C), even though the Sidekick is far weaker than Base (e.g., $52.4\%$ vs.\ $80.3\%$ accuracy on ARC-C). (iii)~In a cross-family pairing, Llama-3.1-8B Base and Qwen2-1.5B Sidekick, Duo improves over B TS in both NLL and ECE on every dataset, with the largest NLL gains of all pairings ($7$--$20\%$), and outperforms the Llama--Llama Duo in NLL and accuracy on all six datasets; as this Sidekick is also larger (1.5B vs.\ 1B), the comparison does not separate model family from Sidekick size. Overall, Duo improves on B TS in NLL in all 24 dataset--pairing combinations and in ECE in 22, changes accuracy by at most $0.4$ points, and is far better calibrated than the twice-as-expensive B DE in every setting. Its gains are therefore not specific to one model pair or family.

Finally, we evaluated the sensitivity of Duo to the amount of validation data used to fit its coefficients. Refitting $(\alpha,\beta)$ on random 20\% and 40\% subsets of the validation set (ten draws each) leaves the coefficients and test NLL nearly unchanged; at 20\%, test NLL is within $0.011$ of its full-data value on five datasets (Tables~\ref{tab:coefficient_sensitivity} and~\ref{tab:nll_sensitivity}). The exception is ARC-C, where 20\% amounts to about 60 examples, and the variance grows. ECE follows the same overall trend but is noisier. These results suggest that a few hundred labeled examples suffice.

\paragraph{FLOP-Performance tradeoff.} The main competitiveness of Duo comes from its computational efficiency. 
Unlike other uncertainty-aware methods that require costly ensembles, repeated stochastic forward passes, or additional Bayesian approximation procedures, Duo achieves similar, and sometimes better, calibration improvements with substantially lower computational overhead. At the same time, it preserves the predictive accuracy of the Base model, making it a practical and scalable alternative for an uncertainty-aware calibration method. On CSQA (Figure~\ref{fig:pareto_csqa}), B BLoB achieves the best NLL and ECE, but Duo also lies on the Pareto frontier at roughly one eighth of its cost; it lies on or next to the frontier on the other datasets as well (Appendix~\ref{app:pareto_frontiers}).

Taken together, these results demonstrate that Asymmetric Duo is flexible across different model configurations and represents a promising approach to improve calibration with only a modest increase in computational cost in this domain.

\subsection{Compatibility with calibration-aware methods}
\label{sec:compat}
\begin{figure}[htbp]
    \centering
    \includegraphics[width=1\linewidth]{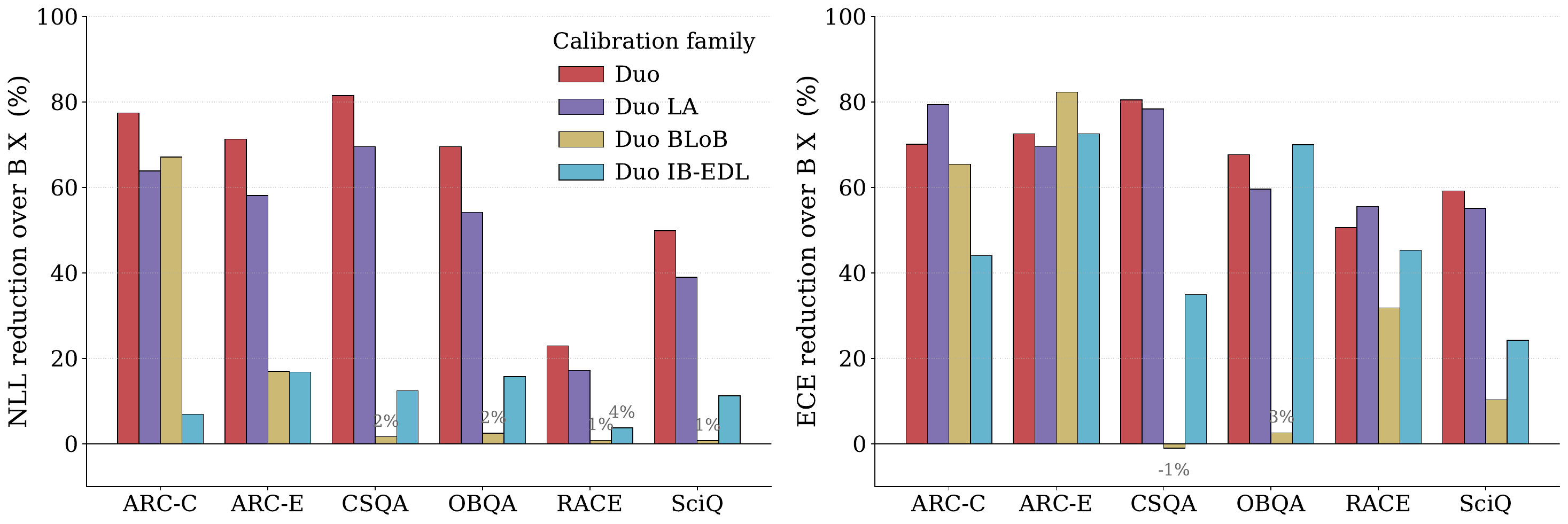}
    \caption{Relative reduction (\%) of NLL (left) and ECE (right) achieved by Duo $X$ over its base predictor B $X$ across the six MCQA benchmarks.}
    \label{fig:duo_compatibility}
\end{figure}

We next ask whether the Duo procedure composes additively with calibration-aware baseline methods. For each calibration family $X \in \{\text{LA},\text{BLoB},\text{IB-EDL}\}$ we form a Duo whose base and sidekick models are calibrated with $X$ and fit $(\alpha,\beta)$ on the validation split, at $\approx 1.22\times$ the inference FLOPs of B $X$. Figure~\ref{fig:duo_compatibility} reports the relative NLL and ECE reduction of Duo $X$ over B $X$. Full numerical results, including standard deviations and accuracy, are deferred to \Cref{tab:compat} in Appendix~\ref{app:compat}. We also exclude a Duo formed from temperature-scaled base and sidekick models, as it reduces to the standard Duo by reparameterization (cf. Section~\ref{sec:duo_interpretation}). The three families behave differently, and the improvements separate into two regimes set by how well-calibrated the underlying base already is.

Laplace-LoRA only mildly recalibrates the Base, so B LA remains overconfident. Duo LA over B LA produces large reductions in NLL and ECE with non-overlapping SD bands in \Cref{tab:compat}, mirroring the Duo-over-Base improvements quantified in Section~\ref{sec:pareto}. In both cases, the optimizer converges to $\alpha,\beta\ll 1$ (Table~\ref{tab:duo_alpha_beta}), so Duo is performing global temperature scaling in addition to combining the two predictors.

On the other hand, BLoB and IB-EDL already calibrate the base model substantially before any fusion, in several cells matching or beating plain Duo, so the room left for Duo to improve is smaller and many cells of \Cref{tab:compat} now have margins within one SD. The optimizer does not compress these base models; it converges to $\alpha\approx 1$ for both families and instead uses $\beta$ to decide how much to incorporate the Sidekick. For Duo BLoB, $\beta$ collapses to (near) zero on RACE and SciQ, so Duo BLoB reduces to B BLoB on those datasets (Figure~\ref{fig:duo_compatibility}). For Duo IB-EDL, $\beta$ stays above zero, so the sidekick still contributes, and the relative gains remain consistent but smaller than those for LA.

\subsection{Ablation: Trivial Duos}
\label{sec:ablation}

The Duo optimization stage is essential. Table~\ref{tab:ablation_trivial} replaces the learned $(\alpha,\beta)$ with the setting, $\alpha=\beta=0.5$ (logit averaging). Unlike the optimized Duo formulation, this variant relies on fixed fusion coefficients and therefore cannot benefit from either global temperature scaling or asymmetric model weighting.

As a result, NLL and ECE degrade markedly on every benchmark: compared to the optimized Duo, trivial averaging increases NLL by $1.2$--$3.2\times$ and ECE by $1.6$--$4.5\times$. Because a common scale cannot change the argmax, it also inherits the accuracy of unweighted fusion, which falls below Base on four of six datasets; only the validation-fitted weights down-weight the weaker Sidekick.

\begin{table}[!t]
\caption{Ablation on the Duo fusion weights. Trivial Duo (Duo ($0.5$)) with $\alpha=\beta=0.5$ (logit averaging) skips the Duo optimization stage. The best method per metric--dataset combination is marked in \textbf{bold}. All metrics are mean $\pm$ SD over 5 repetitions.}

\label{tab:ablation_trivial}
\centering
\footnotesize
\setlength{\tabcolsep}{5pt}
\resizebox{\textwidth}{!}{
\begin{tabular}{llcccccc}
\toprule
Metric & Method & ARC-C & ARC-E & CSQA & OBQA & RACE & SciQ \\
\midrule
\multirow{2}{*}{NLL $\downarrow$}
  & Duo ($0.5$) & $1.442_{\pm0.052}$ & $0.459_{\pm0.042}$ & $1.489_{\pm0.103}$ & $0.523_{\pm0.029}$ & $0.315_{\pm0.008}$ & $0.236_{\pm0.012}$ \\
  & Duo       & $\mathbf{0.469}_{\pm0.009}$ & $\mathbf{0.188}_{\pm0.014}$ & $\mathbf{0.473}_{\pm0.014}$ & $\mathbf{0.302}_{\pm0.007}$ & $\mathbf{0.262}_{\pm0.006}$ & $\mathbf{0.167}_{\pm0.006}$ \\
\midrule
\multirow{2}{*}{Acc $\uparrow$}
  & Duo ($0.5$) & $84.32_{\pm0.84}$ & $93.48_{\pm0.48}$ & $85.11_{\pm0.54}$ & $90.36_{\pm0.36}$ & $91.03_{\pm0.17}$ & $95.14_{\pm0.27}$ \\
  & Duo       & $\mathbf{86.81}_{\pm0.49}$ & $\mathbf{95.14}_{\pm0.40}$ & $\mathbf{85.73}_{\pm0.42}$ & $\mathbf{90.60}_{\pm0.14}$ & $\mathbf{91.82}_{\pm0.14}$ & $\mathbf{95.24}_{\pm0.26}$ \\
\midrule
\multirow{2}{*}{ECE $\downarrow$}
  & Duo ($0.5$) & $11.64_{\pm0.67}$ & $4.30_{\pm0.36}$ & $12.63_{\pm0.36}$ & $6.71_{\pm0.27}$ & $4.46_{\pm0.21}$ & $3.43_{\pm0.43}$ \\
  & Duo       & $\mathbf{3.90}_{\pm1.44}$ & $\mathbf{1.30}_{\pm0.15}$ & $\mathbf{2.81}_{\pm0.89}$ & $\mathbf{3.09}_{\pm0.53}$ & $\mathbf{2.78}_{\pm0.37}$ & $\mathbf{1.94}_{\pm0.42}$ \\
\bottomrule
\end{tabular}}
\end{table}

\section{Mechanistic logit-space analysis of Asymmetric Duos}
\label{sec:logit_distribution}

In this section, we analyze how Duo improves the calibration of LLMs by examining changes in their logit-level behavior. In particular, we compare the logit distributions of the base and sidekick models and Duo. For clarity, we focus on the CSQA dataset using a single seed, while results for the remaining datasets are provided in Figures~\ref{fig:logits_distribution_base_sidekick_arc_c}--\ref{fig:logits_distribution_base_duo_sciq} in Appendix~\ref{app:logit_distributions} and Table~\ref{tab:base_correct_wrong_all} in Appendix~\ref{app:duo_improvement_points}.

Let $\mathcal{D}_{\text{test}}=\{(\mathbf{x}_i,y_i)\}_{i=1}^{N}$ denote the test set, where $\mathbf{x}_i$ is the input and $y_i \in \{1,\dots,K\}$ is the correct class. For each input $\mathbf{x}_i$, the model outputs logits $\mathbf{z}_i = (z_{i,1}, z_{i,2}, \dots, z_{i,K}) \in \mathbb{R}^{K}$, where $z_{i,k}$ denotes the logit assigned to class $k$. The predicted class is $\hat{y}_i = \arg\max_{k \in \{1,\dots,K\}} p_{i,k}$, and the corresponding predictive probabilities are obtained using the softmax function $p_{i,k} = \text{softmax}(\mathbf{z}_i)$.

We consider the following three quantities in our analysis: 
\begin{itemize}[leftmargin=0.5cm]
    \item The logit margin, \(z_{i,y_i} - \max_{k \neq y_i} z_{i,k}\), which measures the gap between the correct-class logit and the highest incorrect-class logit. Positive values indicate that the correct class receives the highest logit, and negative values indicate that at least one incorrect class receives a higher logit; 
    \item confidence, \(p_{i,\hat{y}_i} = \max_{k} p_{i,k}\), which is the probability assigned to the predicted class; 
    \item the correct-class probability, \(p_{i,y_i}\), which is the probability assigned to the correct class.
\end{itemize}

\subsection{Duo softens Base's overconfidence by compressing logit margins} 
\begin{figure}[!htbp]
    \centering
    \includegraphics[width=0.49\linewidth]{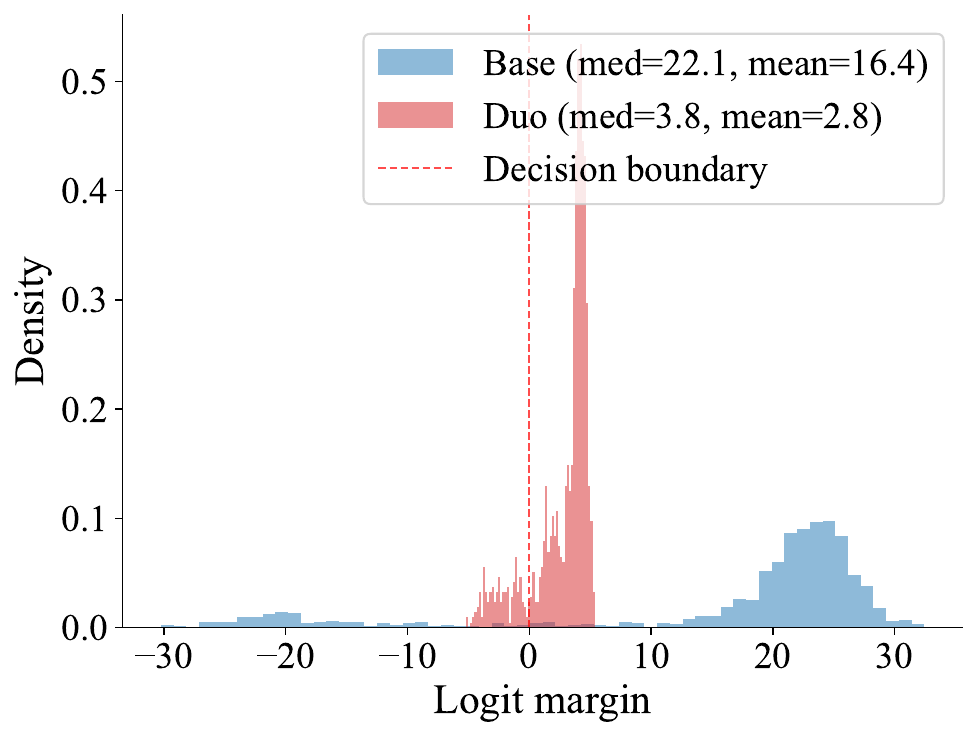}
    \hfill
    \includegraphics[width=0.49\linewidth]{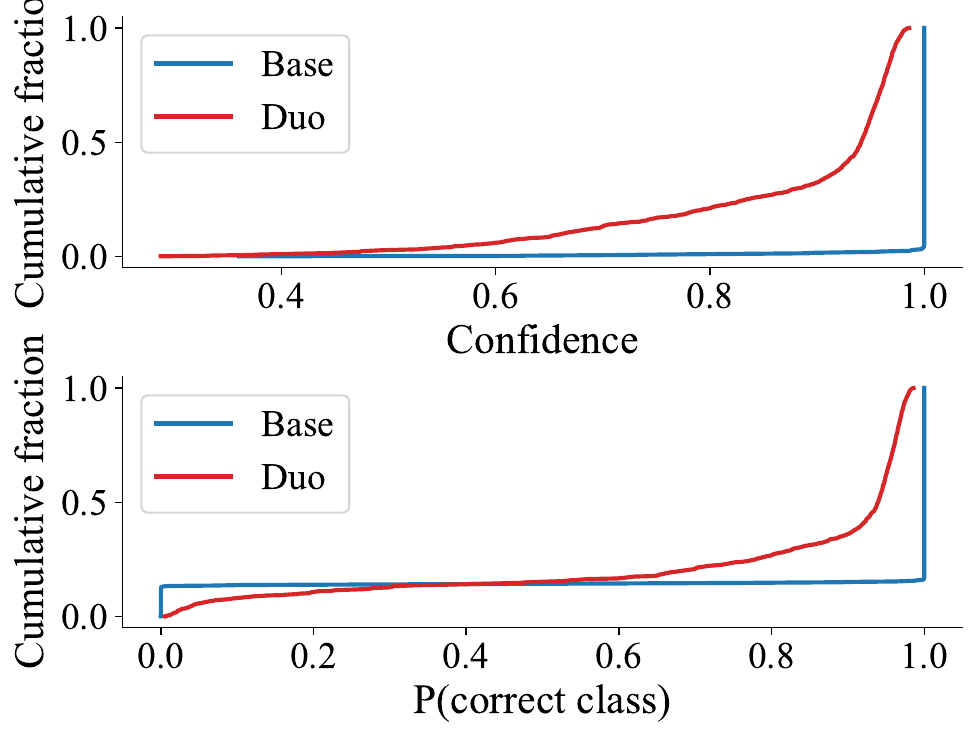}
    
    \caption{Logit margin (left) and empirical cumulative distribution functions (CDFs) of probability and confidence (right) for Duo and Base on the CSQA dataset.}
    \label{fig:logits_distribution_base_duo}
\end{figure}

The confidence CDF in Figure~\ref{fig:logits_distribution_base_duo} (right) shows that the fine-tuned Base places nearly all probability mass on a single option, which explains its poor NLL and ECE despite high accuracy (Table~\ref{tab:results_overall}). Figure~\ref{fig:logits_distribution_base_duo} illustrates how Duo mitigates this overconfidence at the logit level. Duo compresses the logit margin toward zero (left; median $22.1\to3.8$) and yields a more spread-out distribution of confidence and correct-class probability (right). This behavior follows from the unconstrained fusion mechanism: since the weights \(\alpha\) and \(\beta\) are optimized without enforcing a simplex constraint, the resulting logits can be rescaled relative to the original models, leading to a softer predictive distribution. Sections~\ref{sec:disentangling} and~\ref{sec:prediction_change} isolate the part contributed by the Sidekick.

\subsection{Disentangling scaling and asymmetric weighting}
\label{sec:disentangling}

Duo's coefficients control both global logit scaling and the relative contributions of the Base and Sidekick. To separate these effects, we introduce the nested ablation in Table~\ref{tab:scaling_weighting_variants}. Base TS isolates global scaling, symmetric learned fusion adds the Sidekick under an equal-weight constraint, and Duo relaxes this constraint by learning separate coefficients. All variants are fitted on validation NLL using the same optimization procedure. Duo oracle is fitted directly on the test set and serves only as a lower-bound reference.

As shown in Table~\ref{tab:scaling_weighting_nll}, Duo achieves the lowest NLL among the validation-fitted methods and outperforms symmetric learned fusion across all datasets.
Thus, asymmetric weighting provides an additional benefit beyond global scaling and symmetric fusion. Duo also remains within the bounds of Duo oracle, indicating that its coefficients generalize well to the test data.

\newpage

\subsection{Calibration gains come from the wrong predictions of Base} 

\begin{wraptable}{r}{0.5\textwidth}
    \vspace{-2\baselineskip}
    \centering
    \footnotesize
    \setlength{\tabcolsep}{5pt}
    \caption{NLL and ECE of Base and Duo, computed on the CSQA test
    samples that Base classifies correctly/incorrectly.}
    \label{tab:base_correct_wrong}
    \begin{tabular}{ll
      S[table-format=1.3]
      S[table-format=2.3]}
    \toprule
    Metric & Method & {Base Correct} & {Base Wrong} \\
    \midrule
    \multirow{2}{*}{NLL $\downarrow$}
      & Base & {\bfseries 0.003} & {17.705} \\
      & Duo  & 0.121             & {\bfseries \phantom{0}2.552} \\
    \midrule
    \multirow{2}{*}{ECE $\downarrow$}
      & Base & {\bfseries 0.28\phantom{0}} & 98.46\phantom{0} \\
      & Duo  & 9.18\phantom{0}             & {\bfseries 65.59\phantom{0}} \\
    \bottomrule
    \end{tabular}
\end{wraptable}
Table~\ref{tab:base_correct_wrong} splits NLL and ECE by whether Base classifies a test example correctly. A clear asymmetry can be observed. Compressing the margin slightly lowers $p_{i,y_i}$ on examples that Base already classifies correctly (NLL $0.003\to0.121$). The major improvement comes from misclassified samples: under Base, the strongest competing logit dominates by a wide margin, resulting in a significantly small $p_{i,y_i}$ and producing a large NLL contribution ($17.705$) that can outweigh correct samples. Duo's compressed margins increase $p_{i,y_i}$ on these samples, collapsing the per-sample NLL to $2.552$, roughly a seven-fold reduction. At the same time, Duo recovers the argmax of the Base in most cases, so accuracy (Table~\ref{tab:results_overall}) remains largely unchanged or slightly improves. Because NLL is unbounded above, the per-sample improvement on incorrect predictions far exceeds the per-sample degradation on correct ones, and the mean drops sharply. The same Base-Correct/Base-Wrong asymmetry appears in ECE ($0.28 \to 9.18$ versus $98.46 \to 65.59$), and the mechanism is the same redistribution of probability mass: Base concentrates nearly all of its mass in the top-confidence bin, where the gap between $p_{i,\hat{y}_i}$ and empirical bin accuracy is widest. Duo spreads this mass into intermediate-confidence bins whose accuracy more closely matches their stated confidence, shrinking the calibration gap precisely where it dominates the metric.

\subsection{Confidence adjustment versus prediction changes}
\label{sec:prediction_change}
To understand whether Duo's gains arise from changing predictions or adjusting confidence, we partition the test set according to whether Duo changes the Base prediction. Changed predictions are further classified as wrong-to-right or right-to-wrong. We compare Duo with Base TS to exclude improvements attributable solely to temperature scaling. Since temperature scaling preserves the predicted class, the partition is identical for Base and Base TS.

As shown in Table~\ref{tab:prediction_change_ablation}, Duo changes the predicted class on only $0.5\%$--$2.1\%$ of test examples. Nevertheless, examples with unchanged predictions account for $89.1\%$--$103.6\%$ of the total NLL improvement. The share exceeds $100\%$ on RACE because the changed-prediction subset contributes negatively. Wrong-to-right flips generally outnumber right-to-wrong flips, except on RACE, explaining the small and dataset-dependent changes in accuracy. Overall, Duo's calibration gains arise primarily from adjusting confidence while retaining the Base prediction. Because ECE depends on confidence bins, it does not admit the same additive decomposition
as NLL.

\section{Discussion}
\label{sec:discussion}

In this study, we propose and study an adaptation of the Asymmetric Duos framework for uncertainty estimation in LLM classifiers as a simple yet cost-efficient alternative to existing methods. Across six MCQA benchmarks, the Duo framework consistently achieves the best or second-best calibration performance, ranking among the top methods in uncertainty quantification metrics such as ECE and NLL, while maintaining or slightly improving the base model's accuracy. Moreover, the Asymmetric Duo framework achieves these improvements at a fraction of the computational cost of stronger baselines such as BLoB, which often achieves better results but requires substantially higher compute. In contrast, the Duo method lies close to the Pareto frontier, offering a favorable trade-off between calibration quality and efficiency.

Beyond standalone performance, we also show that the Duo framework composes additively with other calibration-aware methods. It provides consistent improvements when applied on top of approaches such as Laplace-LoRA, IB-EDL, and BLoB. This suggests that asymmetric fusion of two distinct models captures complementary information that existing single-model techniques do not exploit.

Overall, these results position Asymmetric Duos as a simple yet very effective and broadly applicable approach to improving uncertainty estimation in LLM classifiers. By combining strong performance, low computational overhead, and compatibility with existing calibration techniques, the method offers a practical alternative to more complex and resource-intensive approaches.

\paragraph{Limitations and future work.} Although our results demonstrate the effectiveness of the Asymmetric Duo framework, several limitations remain. First, we focus on MCQA as a controlled evaluation setting. Extending the analysis to tasks such as open-ended generation or sentiment classification could provide further insights, although defining calibration for open-ended sequences remains challenging. Second, we do not evaluate current state-of-the-art large-scale models. Such experiments could clarify the scalability of the approach but would require substantial computational resources and access to closed-source models. Finally, Duo requires a separate, smaller Sidekick model, which introduces additional memory overhead. Future work could address this limitation by using early exits from the Base model's layers to perform the role of the Sidekick.

\subsection*{AI use statement}

In this work, we used generative AI tools to check grammar and improve the clarity of the content, adjust the presentation of visualizations, and assist with code debugging. We did not use these tools to formulate the research questions or hypotheses, design the methodology or experiments,
generate data, or interpret the results. All AI-assisted text was reviewed and revised by the authors, and all suggested code changes were
manually inspected and tested. The scientific findings and conclusions were determined by the authors. We take responsibility for the final content of this work, including text, claims or artifacts produced with the aid of generative AI.

\subsection*{Ethics statement}

This work is a computational study based on publicly available datasets
and pretrained models. It does not involve human participants or the
collection of personal or sensitive data. We do not identify any
specific ethical concerns beyond the general risks associated with
large language models.

\subsection*{Reproducibility statement}

To support reproducibility, we release the complete codebase at \href{https://github.com/isarustu1-dotcom/asymmetric-llm-duos}{https://github.com/isarustu1-dotcom/asymmetric-llm-duos} , with a README describing how to set up the environment, run the experiments, and reproduce all reported results. All datasets and pretrained models are publicly available; models, datasets, hyperparameters, and training details are listed in Appendix~\ref{app:det_imp}. Every method is evaluated on every dataset with five random seeds.

\newpage
\bibliography{iclr2027_conference}
\bibliographystyle{iclr2027_conference}

\newpage
\appendix

\section{Additional Related Work and Background}
\label{app:additional_background}

\paragraph{Multiple-choice question answering as a testbed for LLM
uncertainty.}

MCQA is a prominent use case of LLMs for evaluating uncertainty since
it reduces free-form generation to a finite decision space. This enables
a meaningful comparison of confidence scores across different models
and methods~\citep{wang2024ubench, kumar2023conformal_llm}. At the same
time, MCQA results can be affected by answer-option ordering, token
selection biases, and the tokenization used for next-token answer
extraction~\citep{pezeshkpour2024order, wei2024selectionbias,
sanzguerrero2025mindthegap}. Despite these limitations, MCQA provides a
suitable framework for controlled comparisons between models and
calibration methods.

\paragraph{Fine-tuning, parameter-efficient adaptation, and
calibration-aware training.}

Fine-tuning is widely used to adapt LLMs to downstream tasks, while
parameter-efficient methods such as Low-Rank Adaptation (LoRA) reduce
the computational cost of this process~\citep{hu2022lora,
ding2023peft}. However, adaptation and calibration serve different
purposes: fine-tuning specializes a model for a task, whereas
calibration improves the reliability of its confidence estimates.
Fine-tuned language models can remain overconfident, motivating
calibration-aware training methods in addition to standard
adaptation~\citep{he2023preserving}. For example,
\citet{li2025calibrating_llm} propose IB-EDL as a calibration-aware
training method for fine-tuned LLMs.

\newpage
\section{Additional figures}
\label{app:appendix_figures}
All figures in this appendix use the main configuration (Qwen2-7B Base, Qwen2-1.5B Sidekick).

\FloatBarrier
\subsection{Pareto frontiers}
\label{app:pareto_frontiers}
\begin{figure}[htbp]
    \centering
    \includegraphics[width=0.80\linewidth]{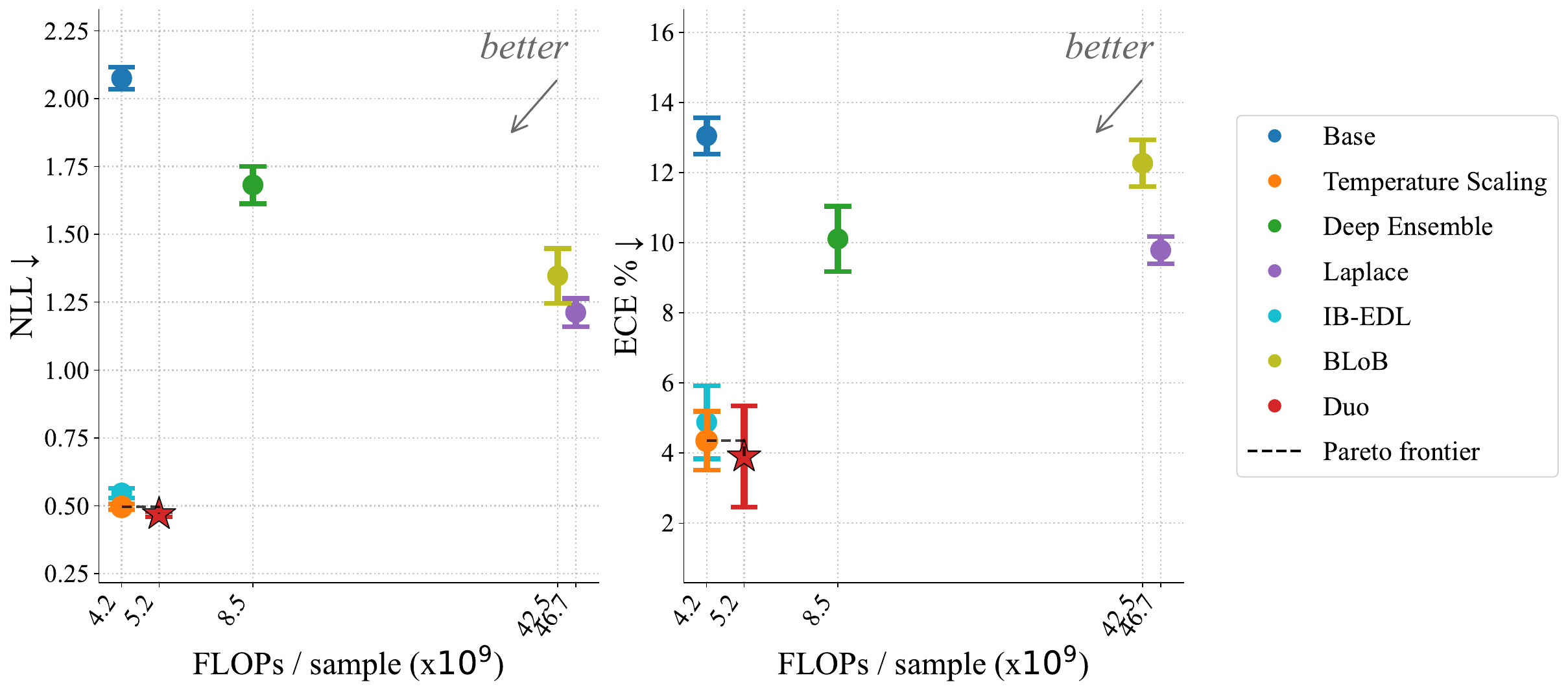}
    \caption{Pareto frontier of FLOPs vs.\ NLL (left) and ECE (right) for ARC-C dataset. For both metrics, Temperature Scaling and Duo lie on Pareto frontier. The method with the best NLL/ECE performance is marked with a star.}
    \label{fig:pareto_arc_c}
\end{figure}

\begin{figure}[htbp]
    \centering
    \includegraphics[width=0.80\linewidth]{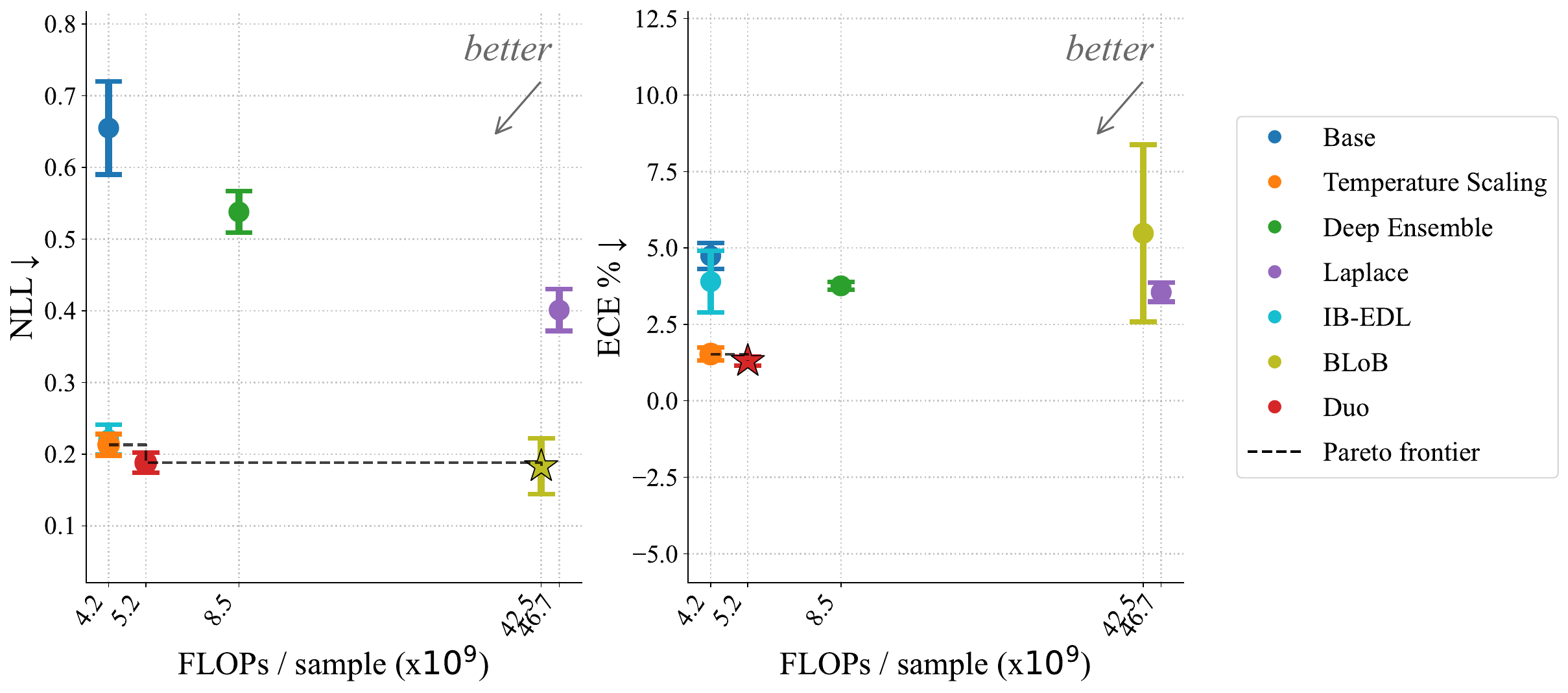}
    \caption{Pareto frontier of FLOPs vs.\ NLL (left) and ECE (right) for the ARC-E dataset. For NLL, Temperature Scaling, Duo, and BLoB lie on the Pareto frontier; for ECE, Temperature Scaling and Duo lie on the Pareto frontier. The method with the best NLL/ECE performance is marked with a star.}
    \label{fig:pareto_arc_e}
\end{figure}

\begin{figure}[htbp] 
    \centering 
    \includegraphics[width=0.80\linewidth]{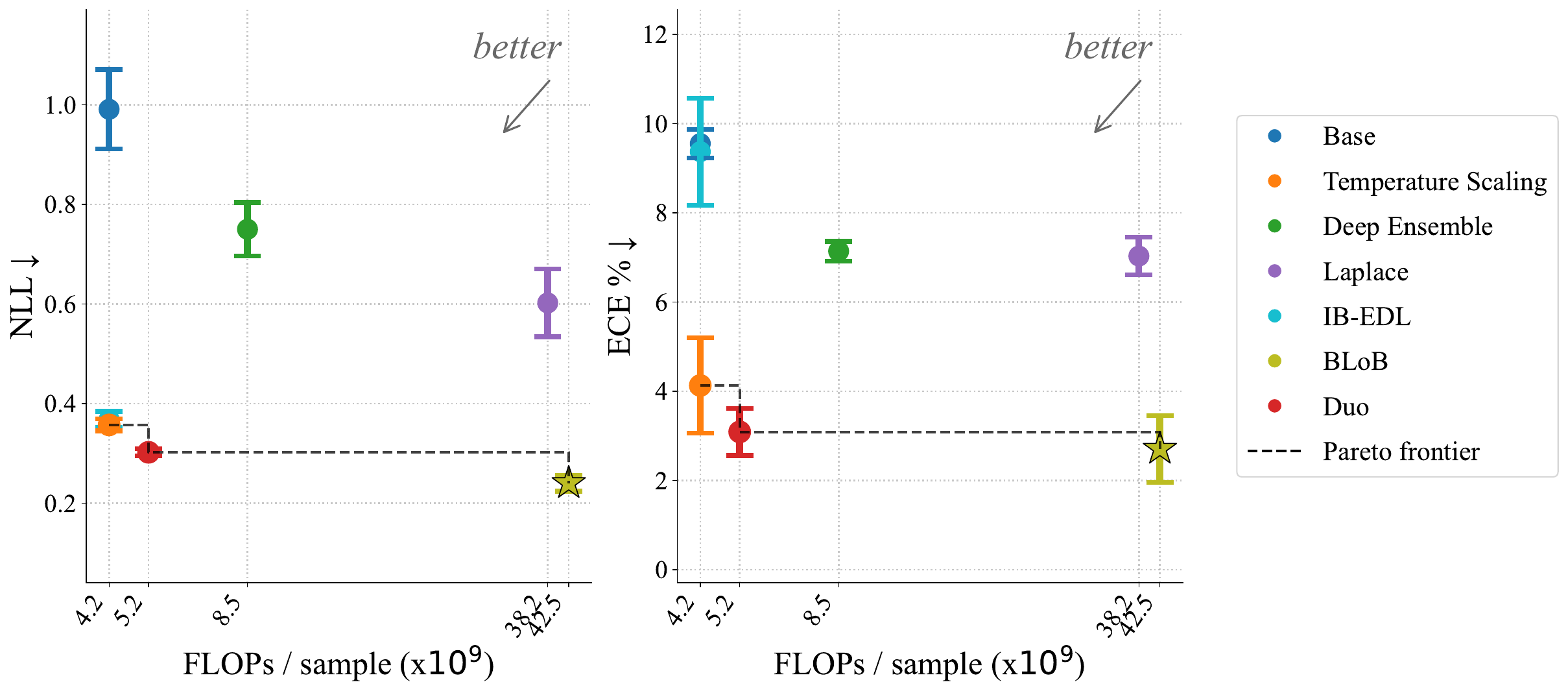} 
    \caption{Pareto frontier of FLOPs vs.\ NLL (left) and ECE (right) for OBQA dataset. For both metrics, Temperature Scaling, Duo, and BLoB lie on Pareto frontier. The method with the best NLL/ECE performance is marked with a star.} 
    \label{fig:pareto_obqa} 
\end{figure}

\begin{figure}[htbp] 
    \centering 
    \includegraphics[width=0.80\linewidth]{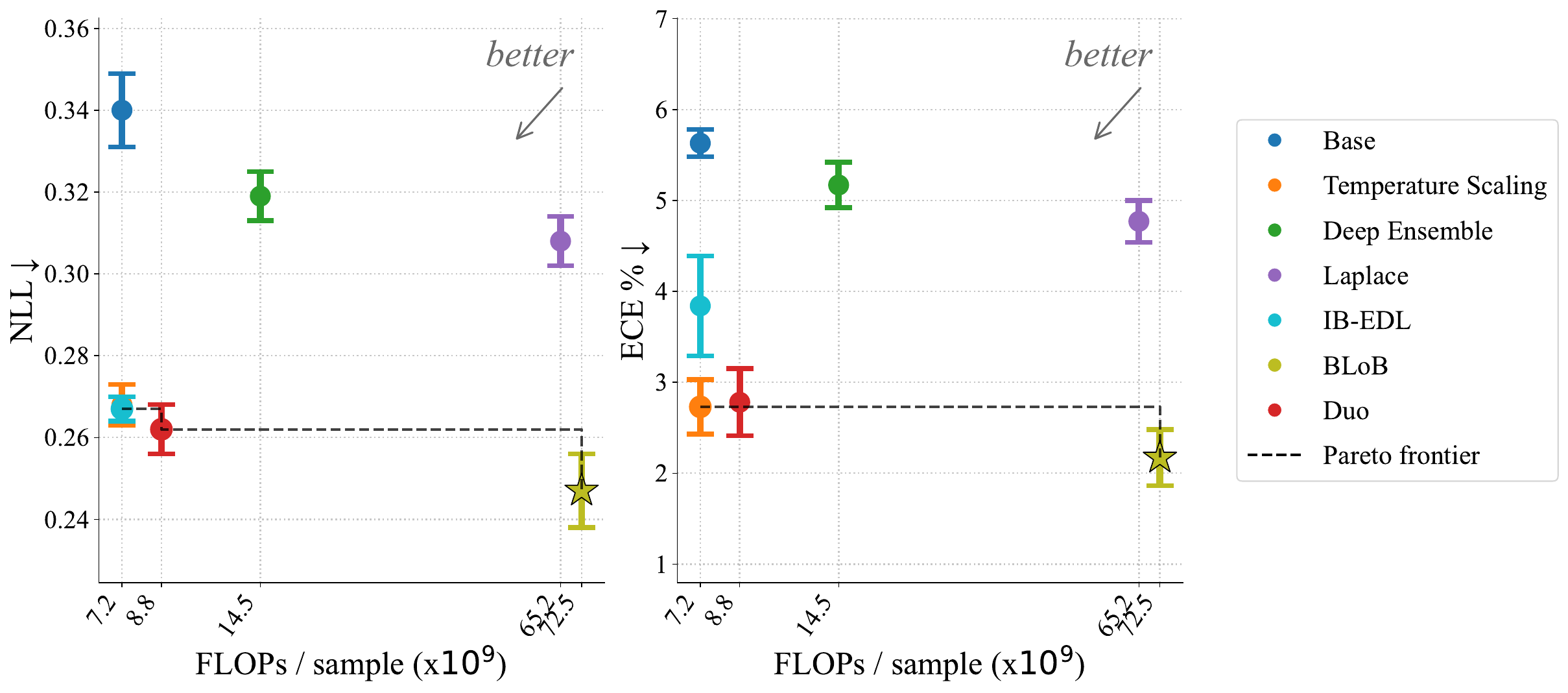} 
    \caption{Pareto frontier of FLOPs vs.\ NLL (left) and ECE (right) for RACE dataset. For NLL, IB-EDL, Duo, and BLoB lie on the Pareto frontier; for ECE, Temperature Scaling and Duo lie on the Pareto frontier. The method with the best NLL/ECE performance is marked with a star.} 
    \label{fig:pareto_race} 
\end{figure}

\begin{figure}[htbp] 
    \centering 
    \includegraphics[width=0.80\linewidth]{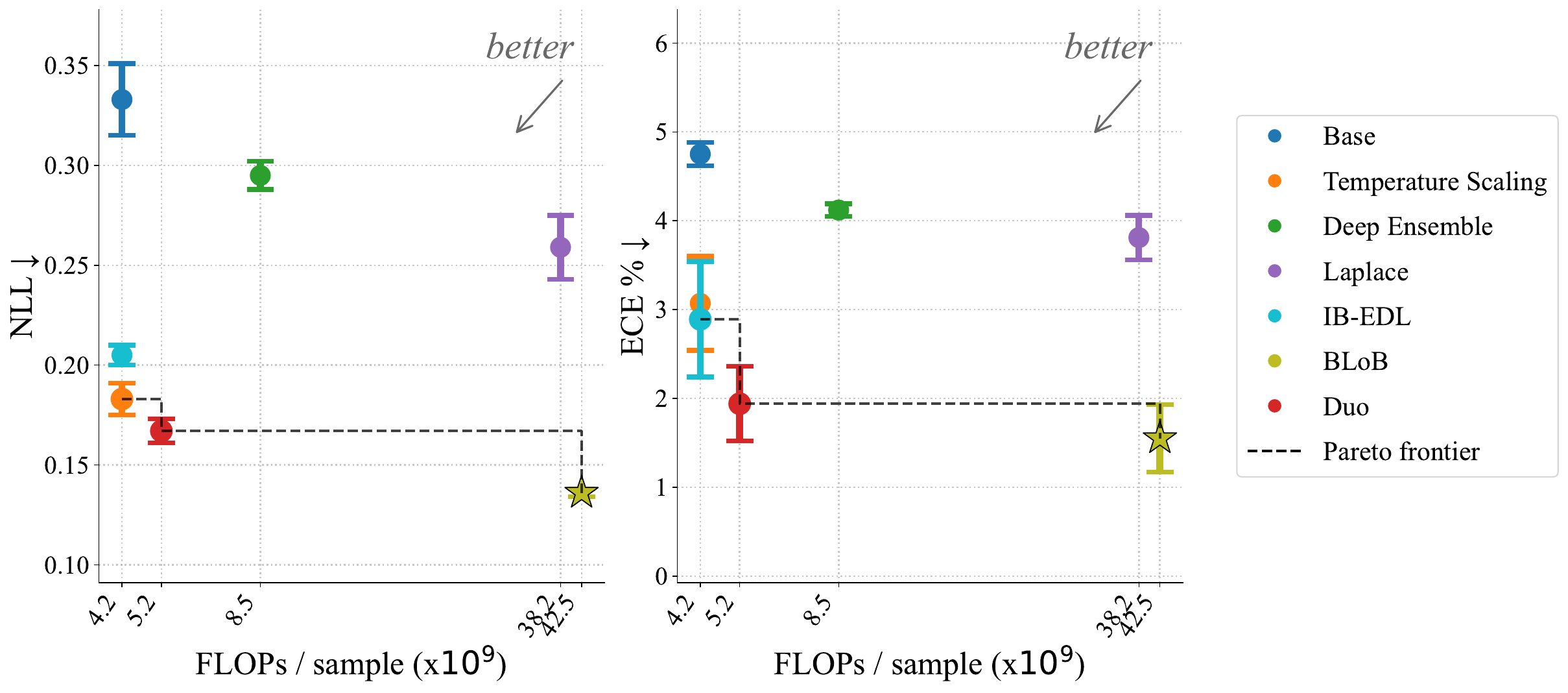} 
    \caption{Pareto frontier of FLOPs vs.\ NLL (left) and ECE (right) for SciQ dataset. For NLL, Temperature Scaling, Duo, and BLoB lie on the Pareto frontier; for ECE, IB-EDL, Duo, and BLoB lie on the Pareto frontier. The method with the best NLL/ECE performance is marked with a star.} 
    \label{fig:pareto_sciq} 
\end{figure}

The per-dataset Pareto frontiers in Figures~\ref{fig:pareto_arc_c}--\ref{fig:pareto_sciq} confirm that the cost-calibration story discussed for CSQA in the main text is not dataset-specific. Across all six benchmarks, Duo sits in the lower-left ``elbow'' alongside B TS, achieving comparable NLL and ECE at $\approx 1.22\times$ the FLOPs of the base model and at $\approx1/8$ the FLOPs of B BLoB. The one partial exception is RACE, where B BLoB and B TS edge Duo on ECE by a small margin (within Duo's standard-deviation band), so Duo is no longer strictly Pareto-dominant there but still lies on the low-cost segment of the frontier.

\newpage

\FloatBarrier

\newpage
\subsection{Logit distributions}
\label{app:logit_distributions}
\begin{figure}[!htbp]
    \centering
    
    \includegraphics[width=0.48\linewidth]{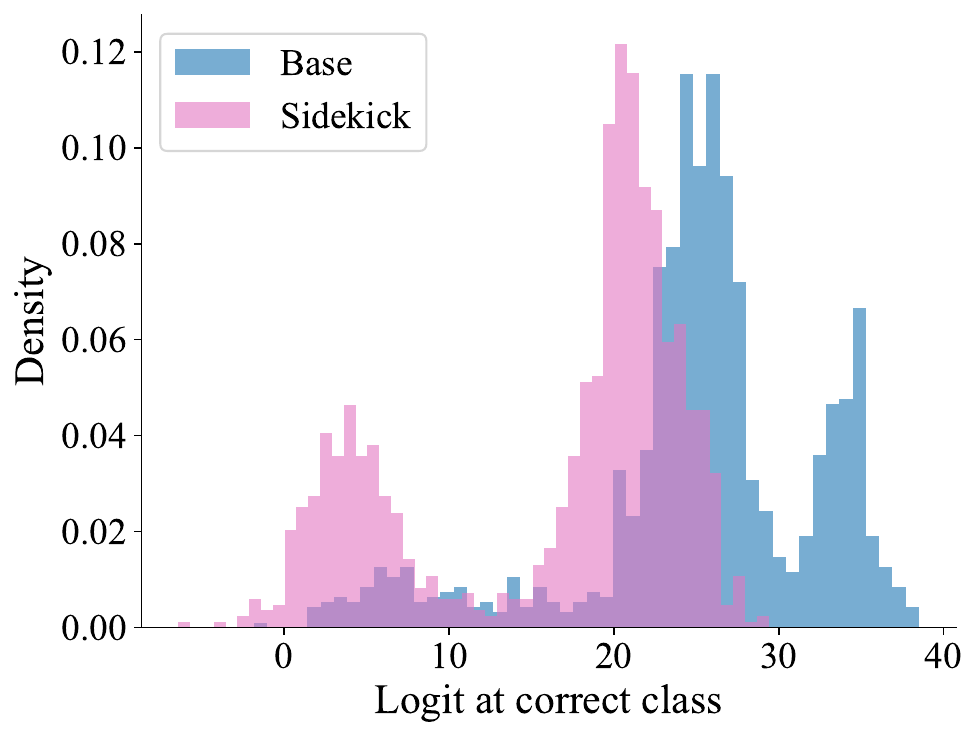}
    \hfill
    \includegraphics[width=0.48\linewidth]{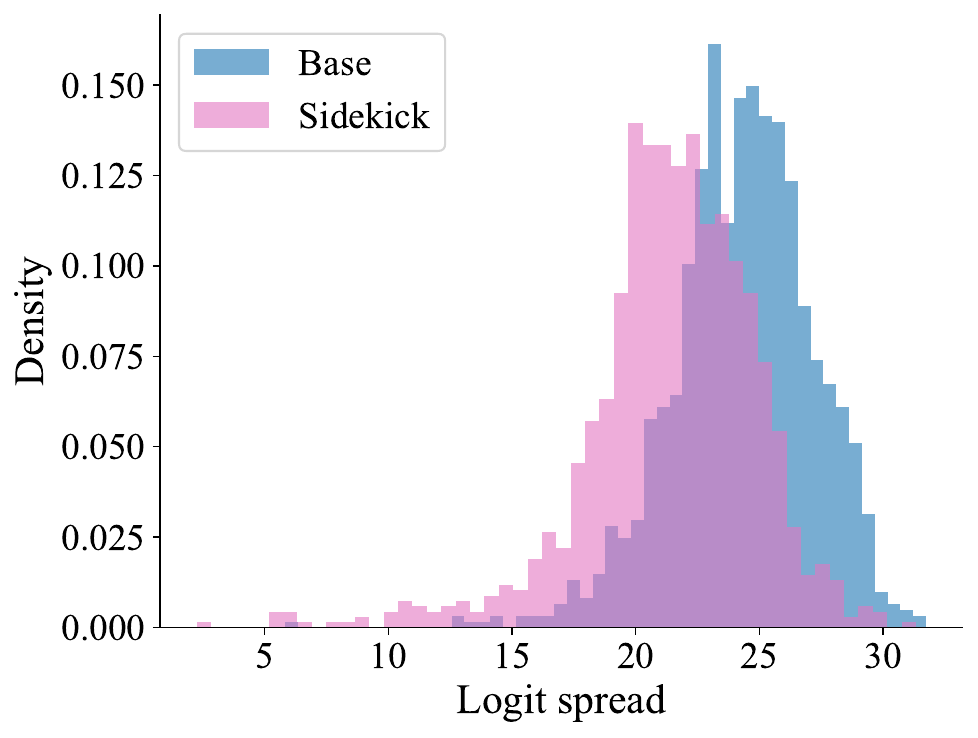}
    
    \caption{Logit (left) and logit spread (right) distributions of base and sidekick models for ARC-C dataset.}
    \label{fig:logits_distribution_base_sidekick_arc_c}
\end{figure}

\begin{figure}[!htbp]
    \centering
    
    \includegraphics[width=0.48\linewidth]{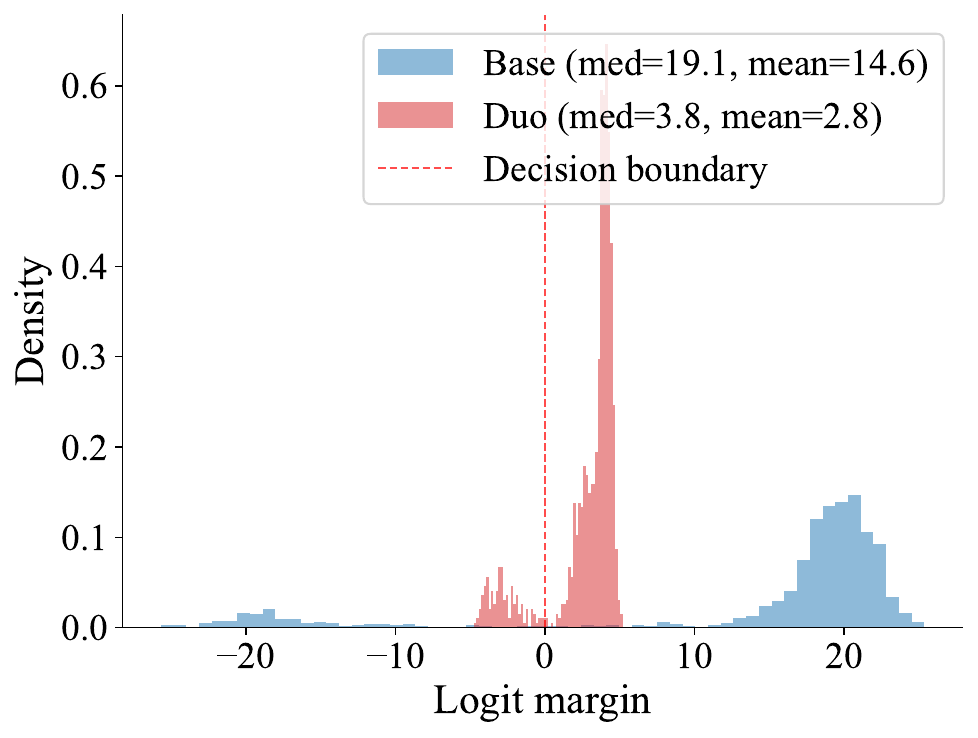}
    \hfill
    \includegraphics[width=0.48\linewidth]{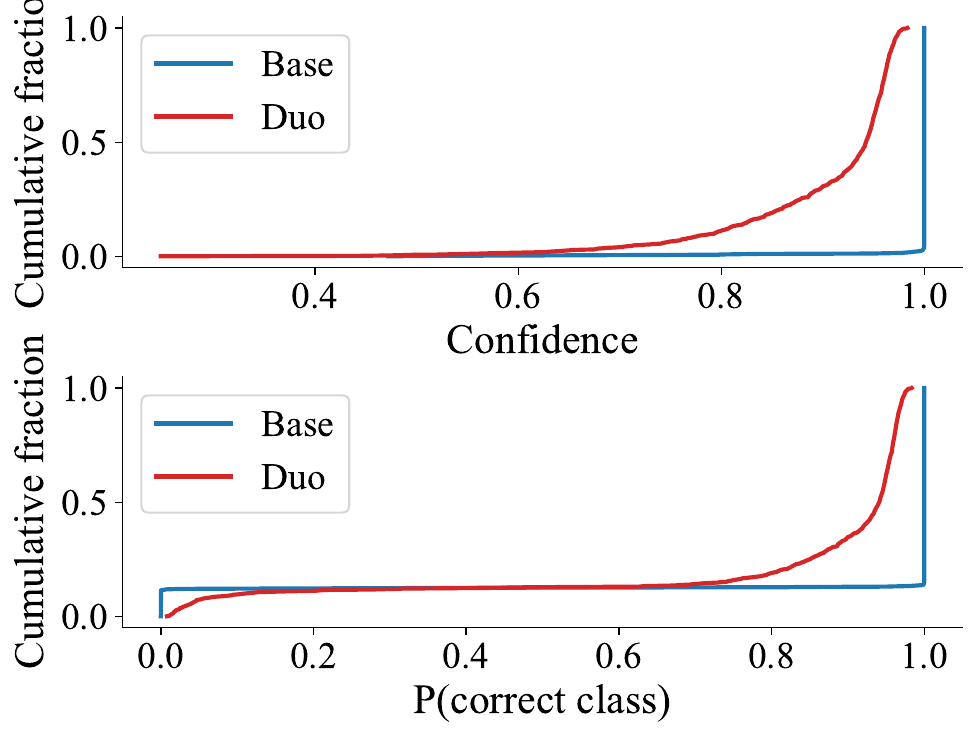}
    
    \caption{Logit margin (left) and empirical cumulative distribution functions (CDF) of probability and confidence (right) for Duo and Base for ARC-C dataset.}
    \label{fig:logits_distribution_base_duo_arc_c}
\end{figure}

\begin{figure}[!htbp]
    \centering
    
    \includegraphics[width=0.48\linewidth]{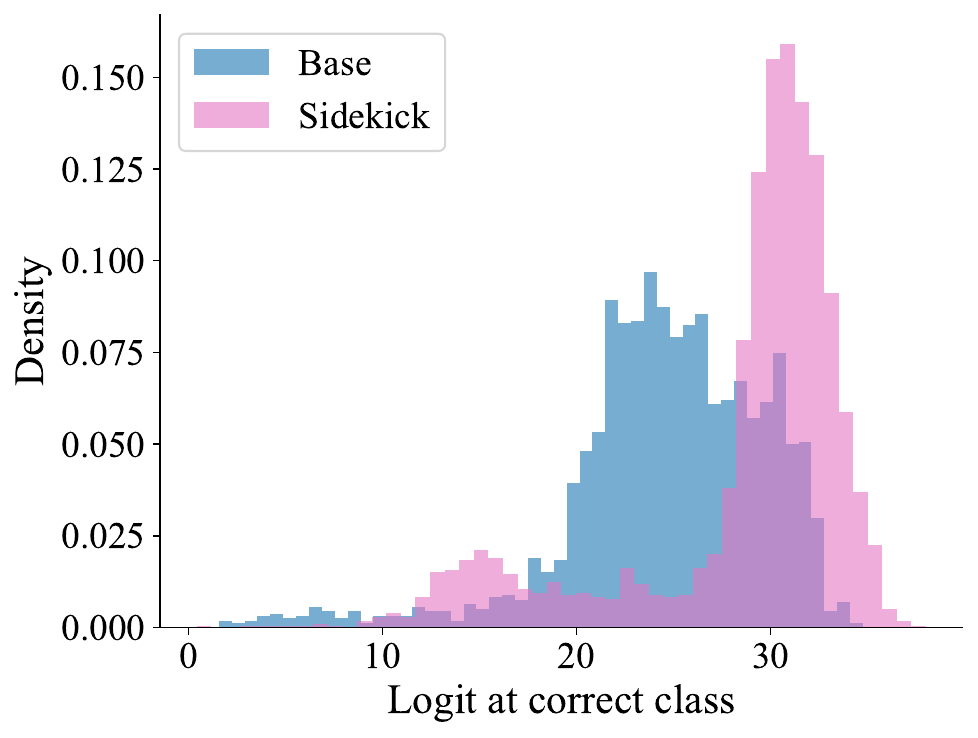}
    \hfill
    \includegraphics[width=0.48\linewidth]{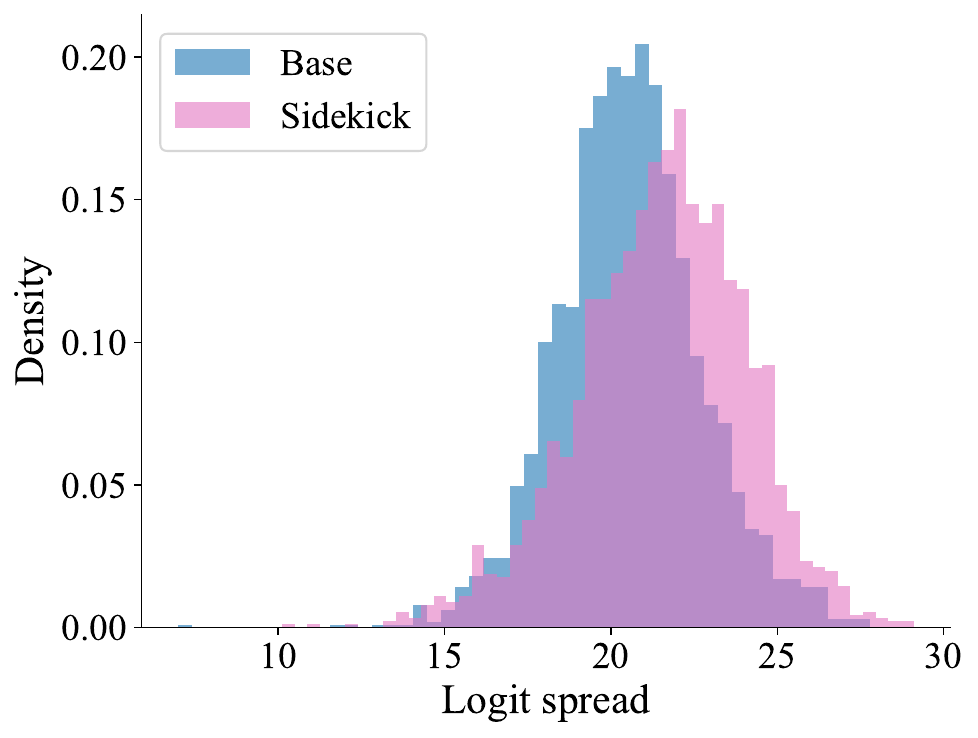}
    
    \caption{Logit (left) and logit spread (right) distributions of base and sidekick models for ARC-E dataset.}
    \label{fig:logits_distribution_base_sidekick_arc_e}
\end{figure}

\begin{figure}[!htbp]
    \centering
    
    \includegraphics[width=0.48\linewidth]{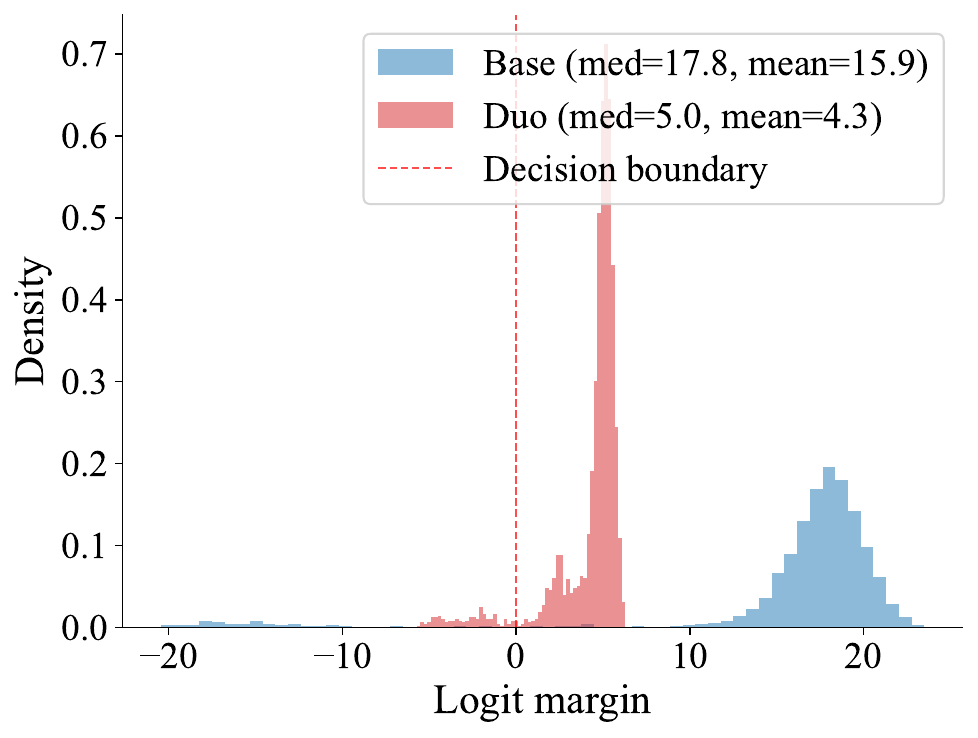}
    \hfill
    \includegraphics[width=0.48\linewidth]{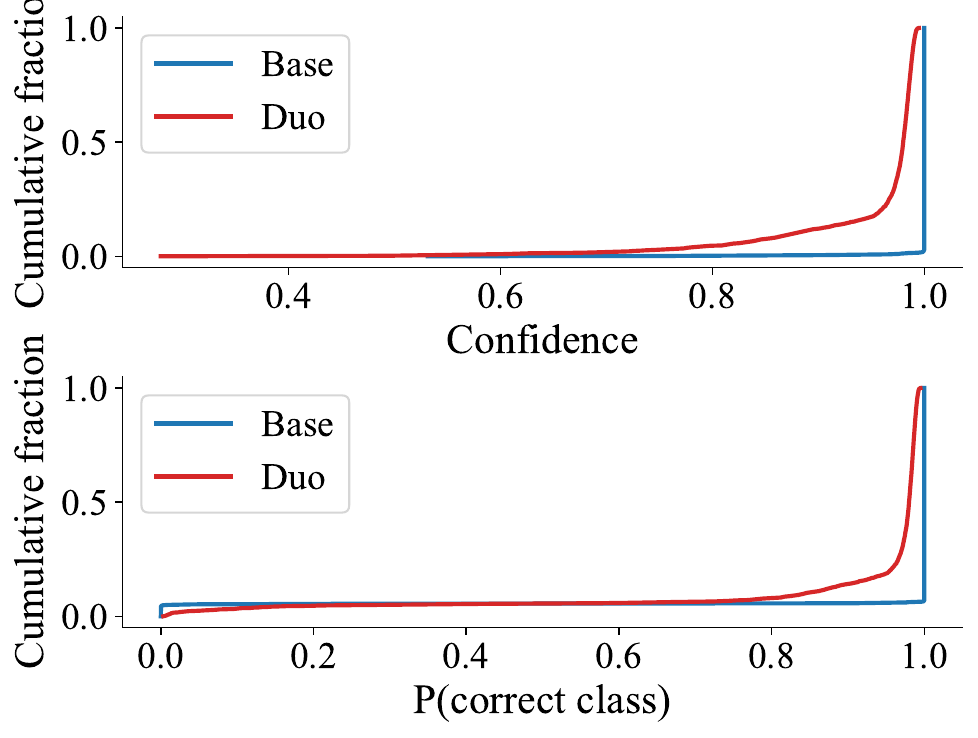}
    
    \caption{Logit margin (left) and empirical cumulative distribution functions (CDF) of probability and confidence (right) for Duo and Base for ARC-E dataset.}
    \label{fig:logits_distribution_base_duo_arc_e}
\end{figure}

\begin{figure}[!htbp]
    \centering
    
    \includegraphics[width=0.48\linewidth]{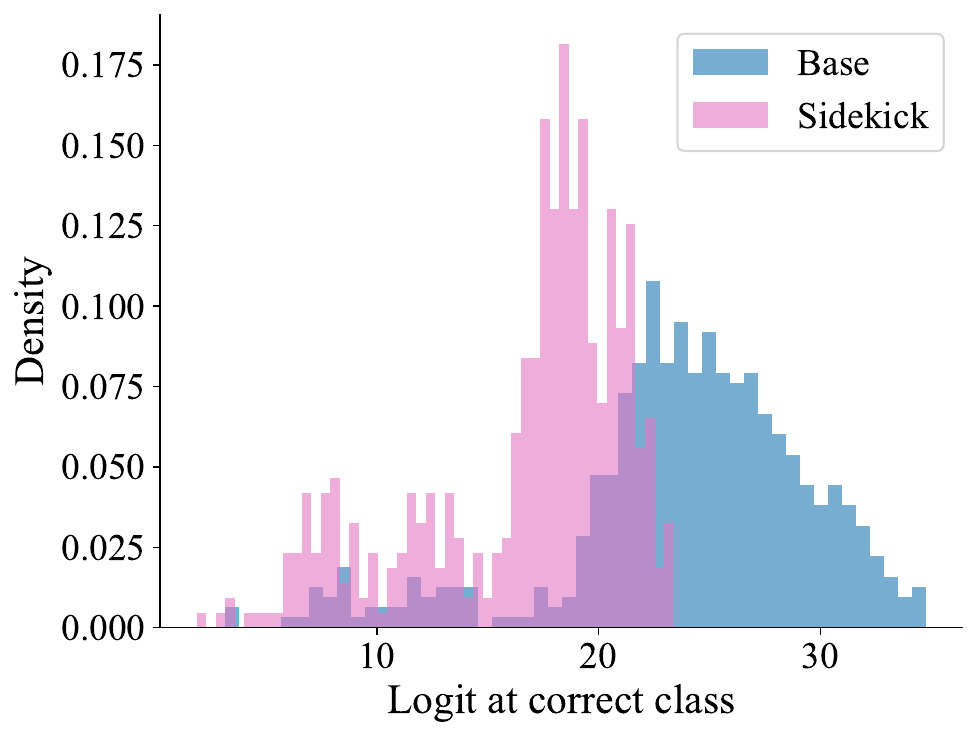}
    \hfill
    \includegraphics[width=0.48\linewidth]{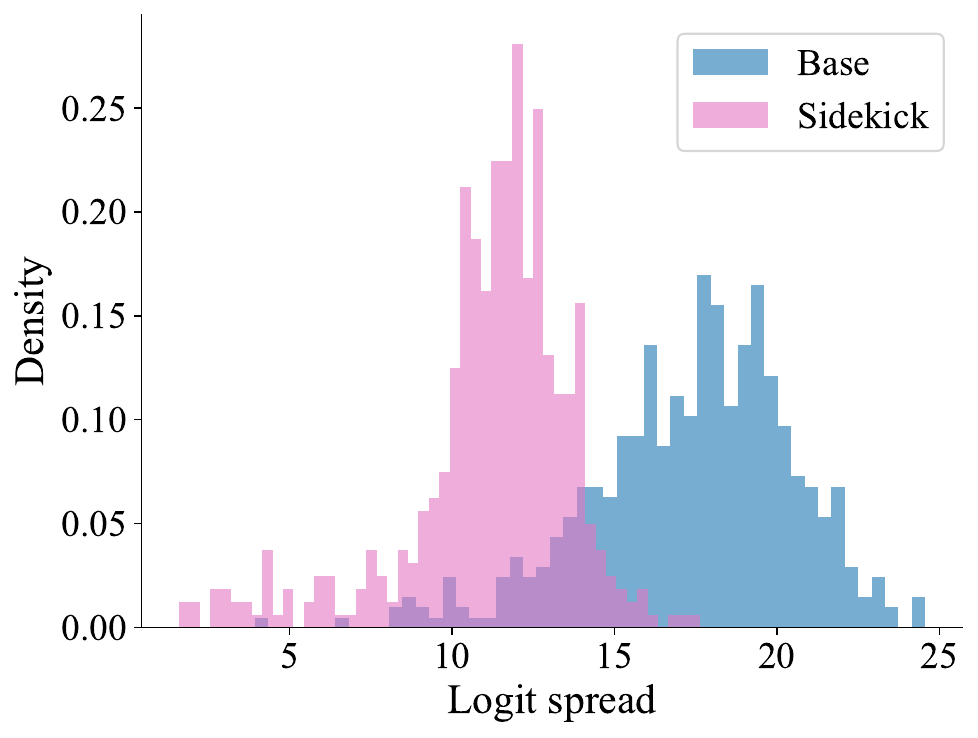}
    
    \caption{Logit (left) and logit spread (right) distributions of base and sidekick models for OBQA dataset.}
    \label{fig:logits_distribution_base_sidekick_obqa}
\end{figure}

\begin{figure}[!htbp]
    \centering
    
    \includegraphics[width=0.48\linewidth]{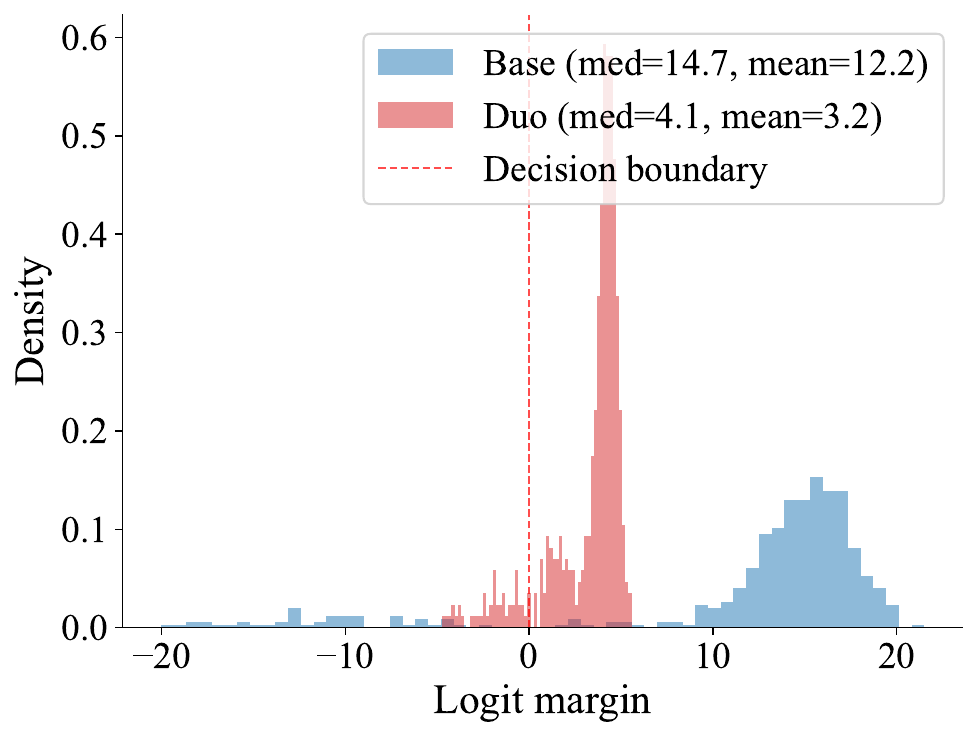}
    \hfill
    \includegraphics[width=0.48\linewidth]{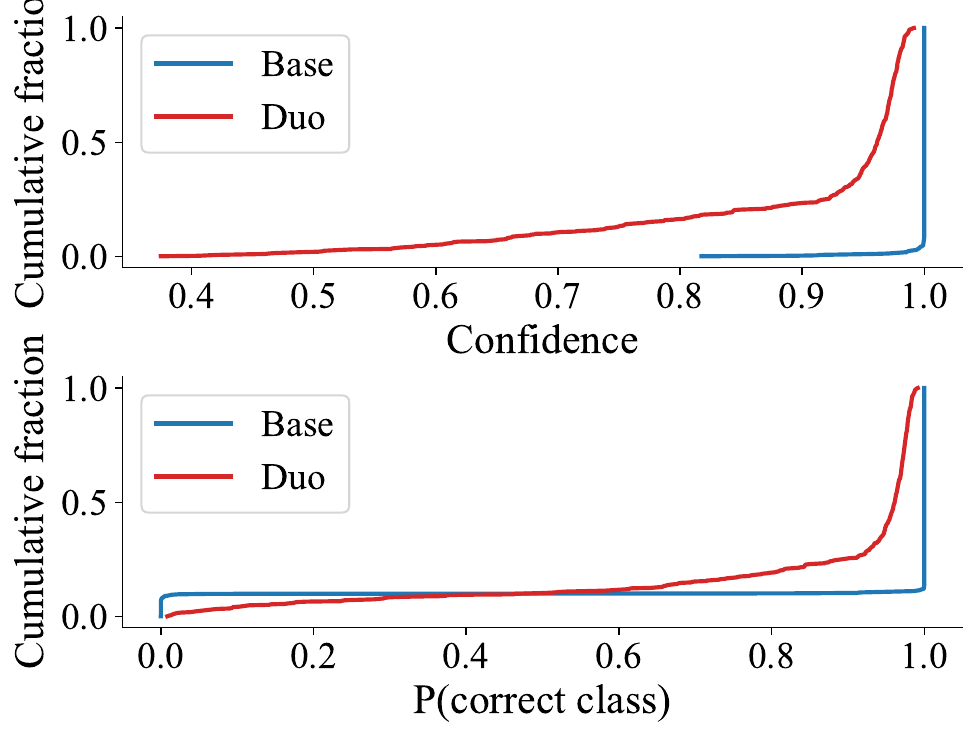}
    
    \caption{Logit margin (left) and empirical cumulative distribution functions (CDF) of probability and confidence (right) for Duo and Base for OBQA dataset.}
    \label{fig:logits_distribution_base_duo_obqa}
\end{figure}

\begin{figure}[!htbp]
    \centering
    
    \includegraphics[width=0.48\linewidth]{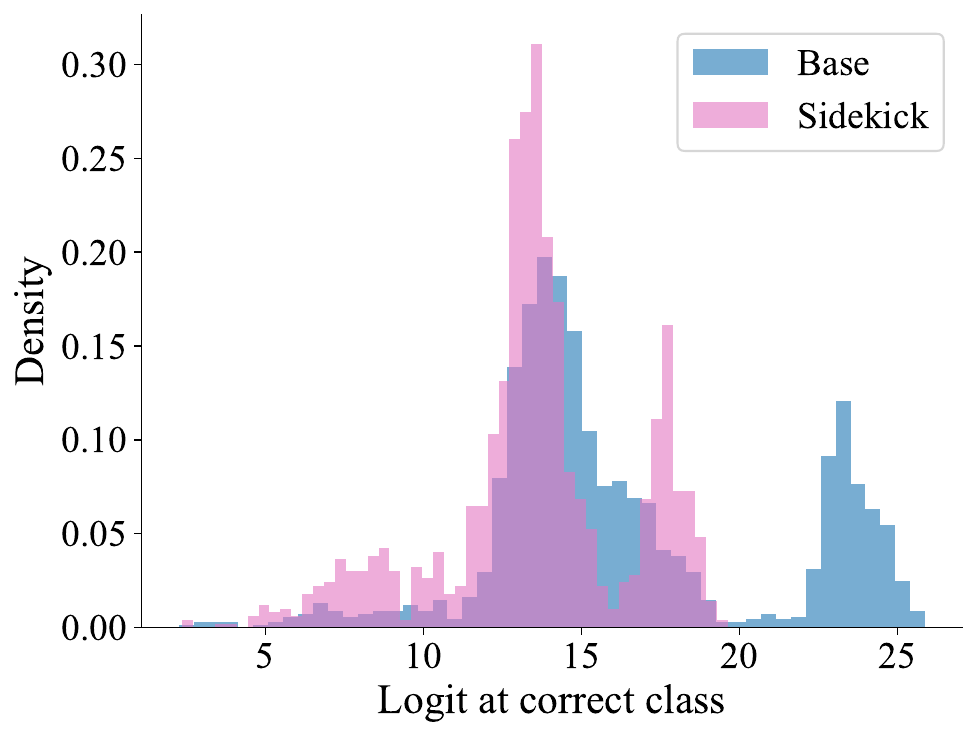}
    \hfill
    \includegraphics[width=0.48\linewidth]{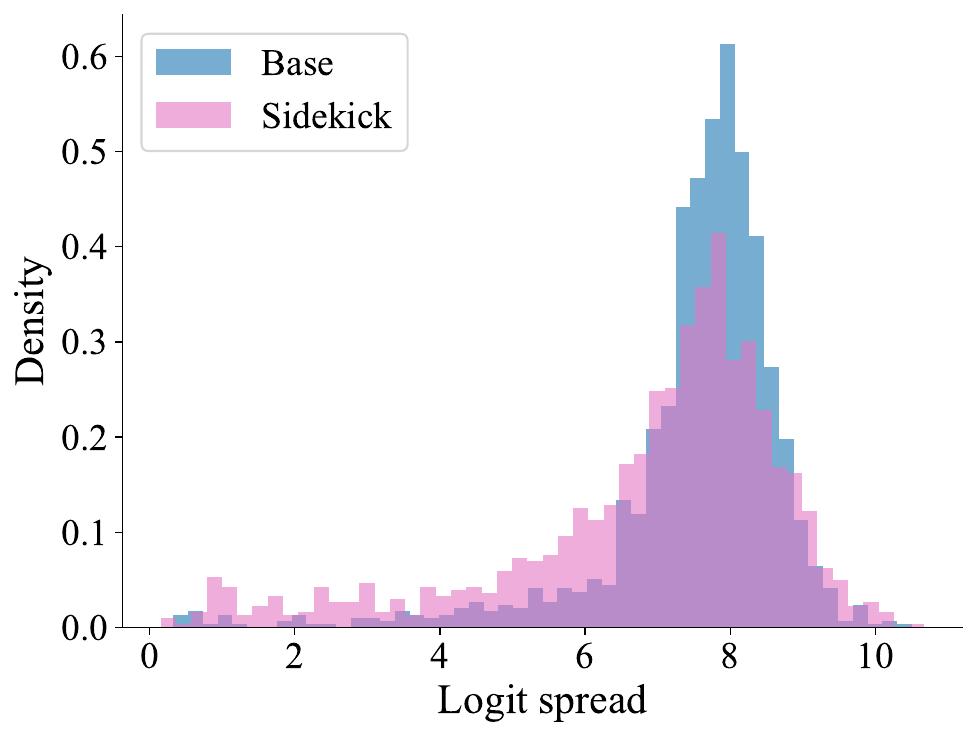}
    
    \caption{Logit (left) and logit spread (right) distributions of base and sidekick models for RACE dataset.}
    \label{fig:logits_distribution_base_sidekick_race}
\end{figure}

\begin{figure}[!htbp]
    \centering
    
    \includegraphics[width=0.48\linewidth]{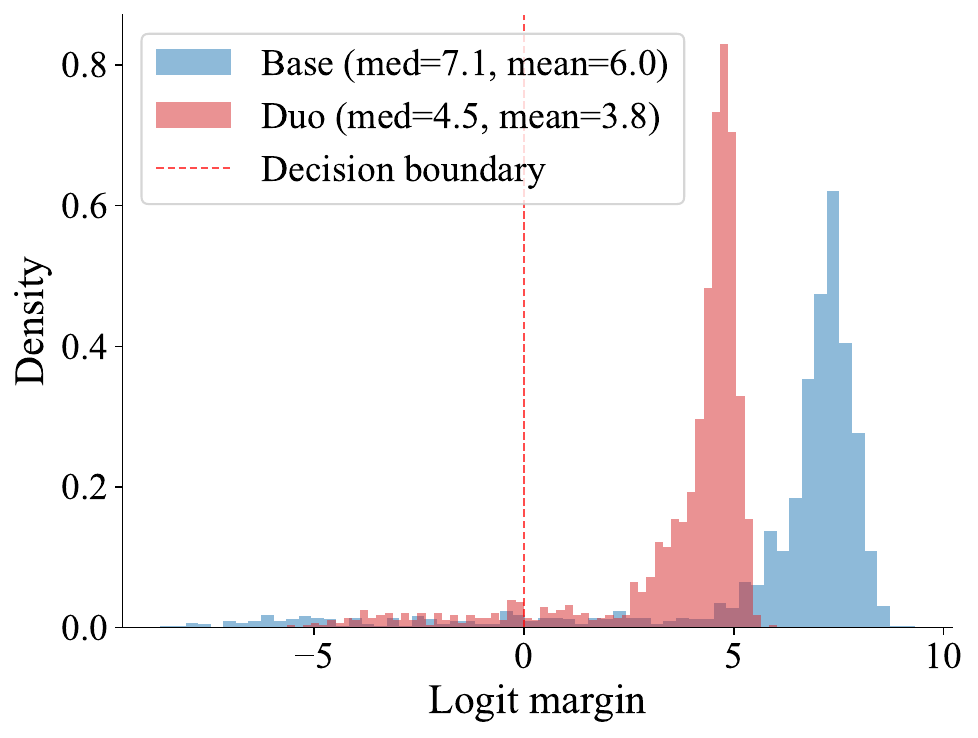}
    \hfill
    \includegraphics[width=0.48\linewidth]{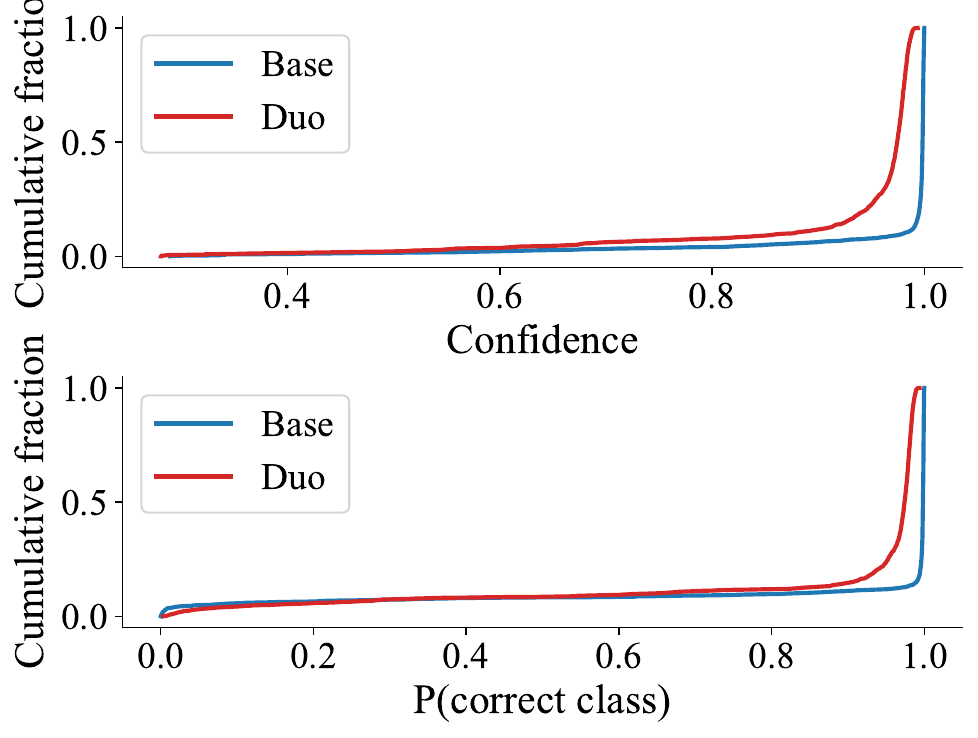}
    
    \caption{Logit margin (left) and empirical cumulative distribution functions (CDF) of probability and confidence (right) for Duo and Base for RACE dataset.}
    \label{fig:logits_distribution_base_duo_race}
\end{figure}

\begin{figure}[!htbp]
    \centering
    
    \includegraphics[width=0.48\linewidth]{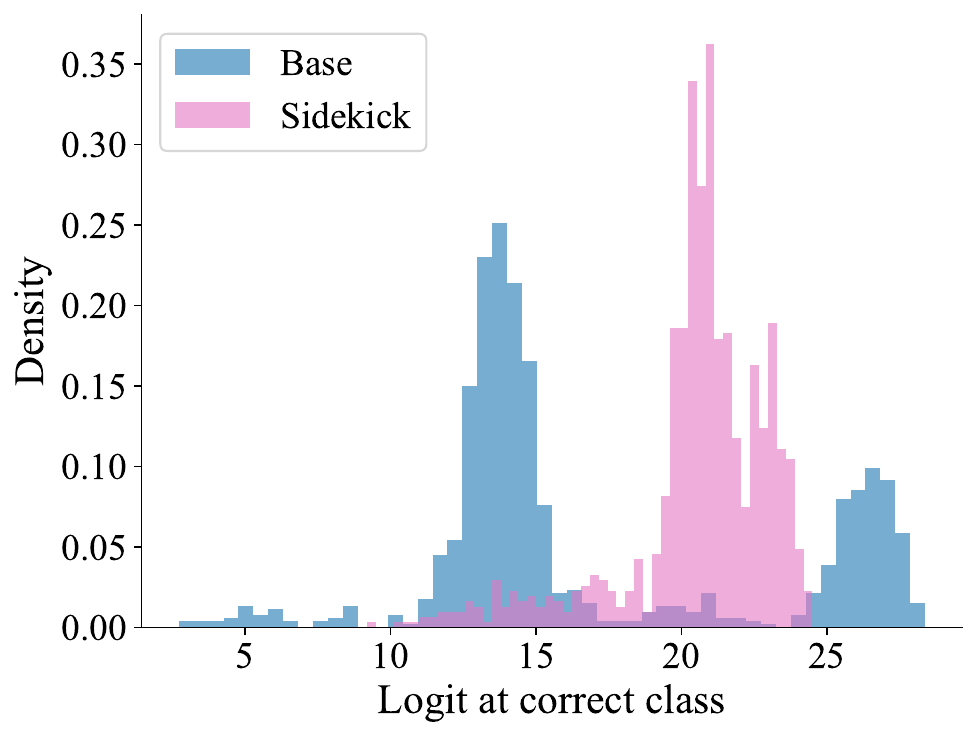}
    \hfill
    \includegraphics[width=0.48\linewidth]{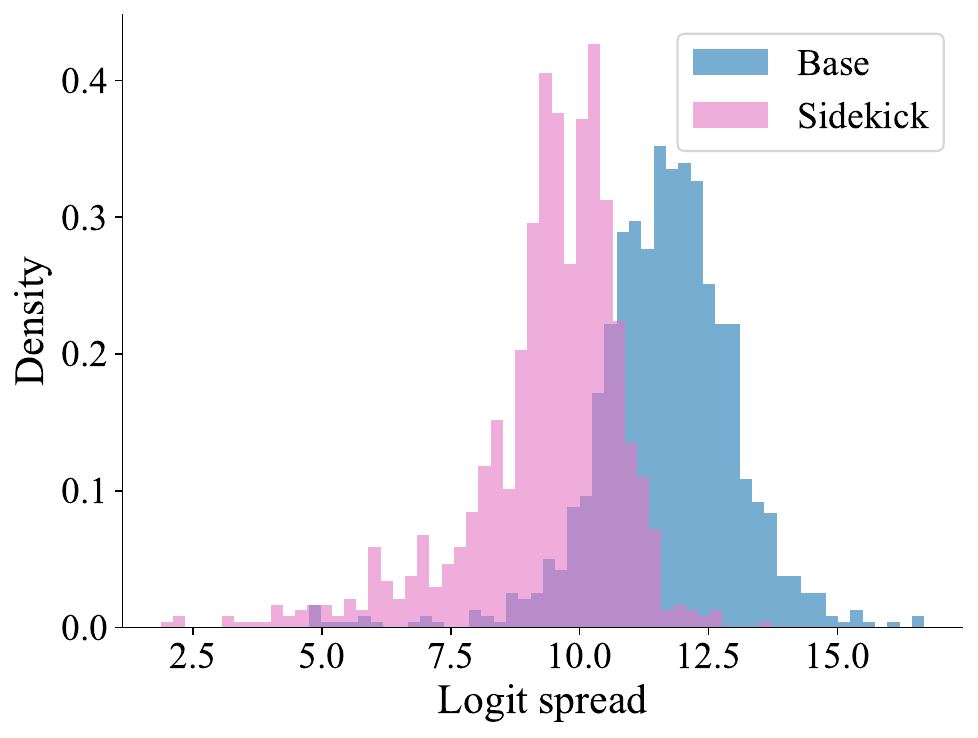}
    
    \caption{Logit (left) and logit spread (right) distributions of base and sidekick models for SciQ dataset.}
    \label{fig:logits_distribution_base_sidekick_sciq}
\end{figure}

\begin{figure}[!htbp]
    \centering
    
    \includegraphics[width=0.48\linewidth]{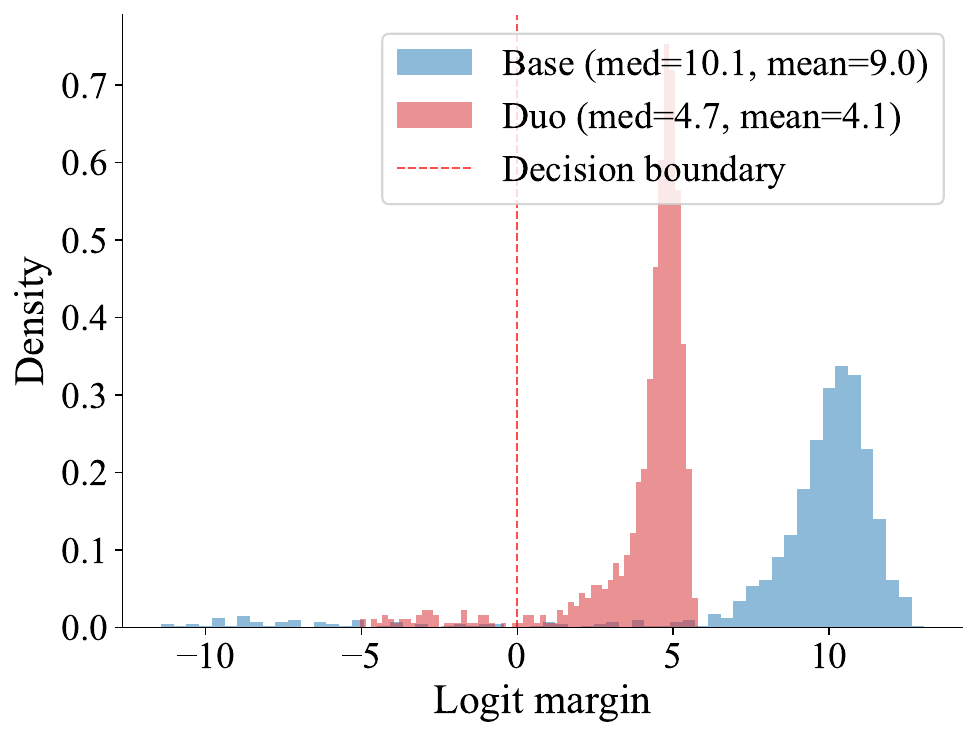}
    \hfill
    \includegraphics[width=0.48\linewidth]{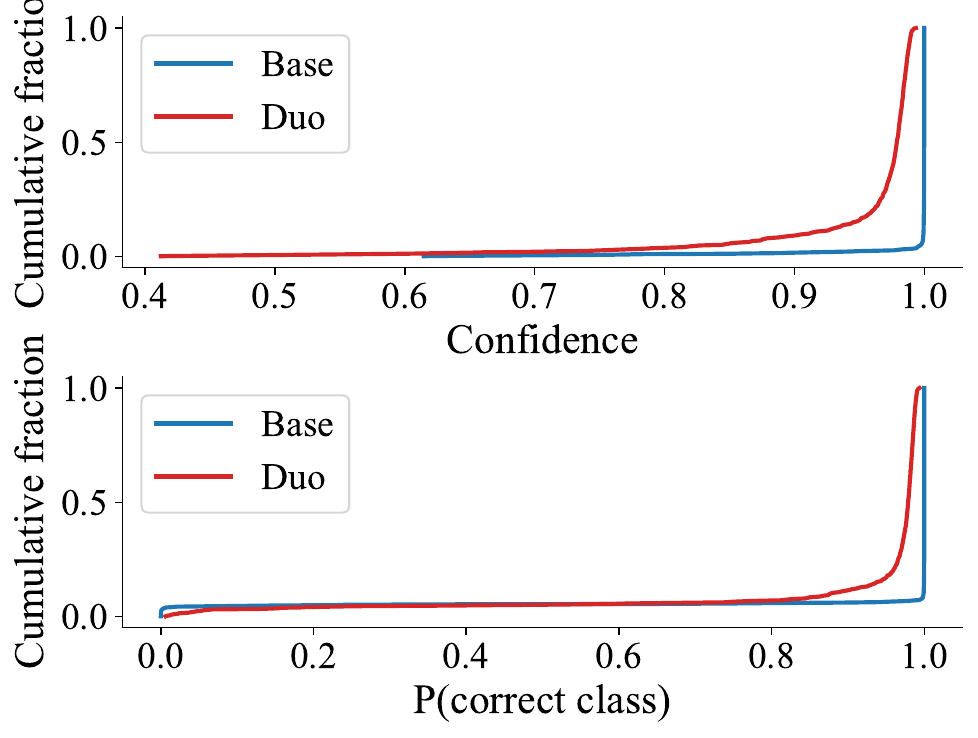}
    
    \caption{Logit margin (left) and empirical cumulative distribution functions (CDF) of probability and confidence (right) for Duo and Base for SciQ dataset.}
    \label{fig:logits_distribution_base_duo_sciq}
\end{figure}

The per-dataset logit/spread/margin/CDF plots in Figures~\ref{fig:logits_distribution_base_sidekick_arc_c}--\ref{fig:logits_distribution_base_duo_sciq} show the same logit-space mechanism we identified for CSQA in Section~\ref{sec:logit_distribution} on every other benchmark. Base and sidekick models concentrate correct-class logits at large positive values with a wide spread, and the Duo predictor systematically pulls the logit margin toward zero. We read this as evidence that asymmetric logit fusion mitigates overconfidence through a single, dataset-agnostic mechanism. The considered quantities are the correct-class logit, \(z_{i,y_i}\), which denotes the logit assigned to the correct class; the logit spread, \(\max_{k} z_{i,k} - \min_{k} z_{i,k}\), which measures the range of the output logits; as well as the logit margin, confidence, and correct-class probability described in \cref{sec:logit_distribution}.

\begin{figure}[htbp]
    \centering
    
    \includegraphics[width=0.49\linewidth]{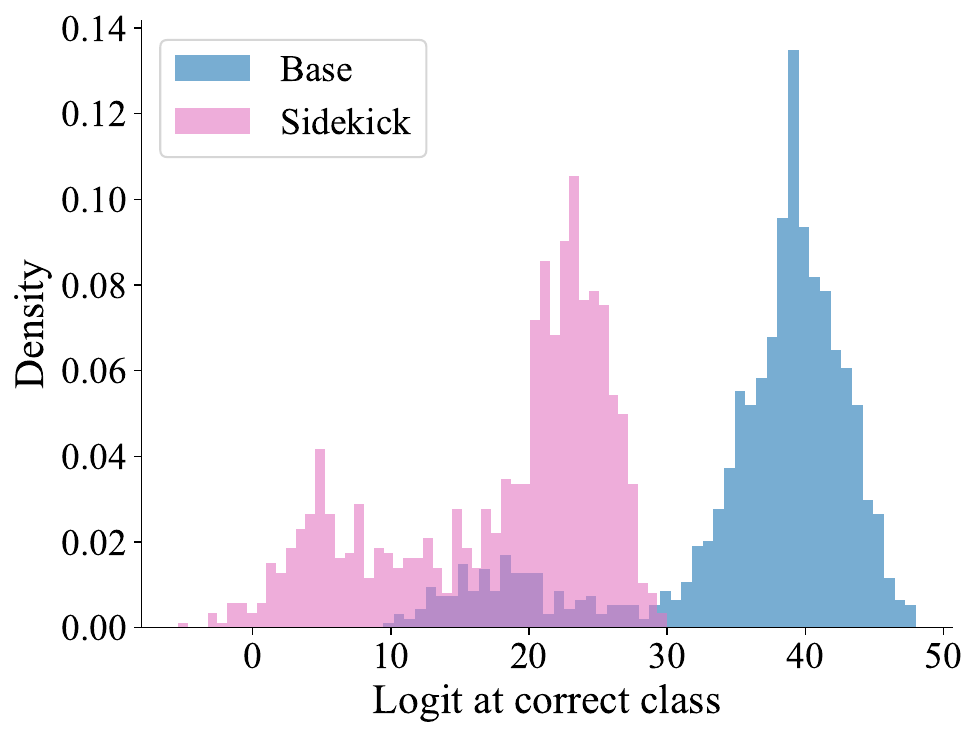}
    \hfill
    \includegraphics[width=0.49\linewidth]{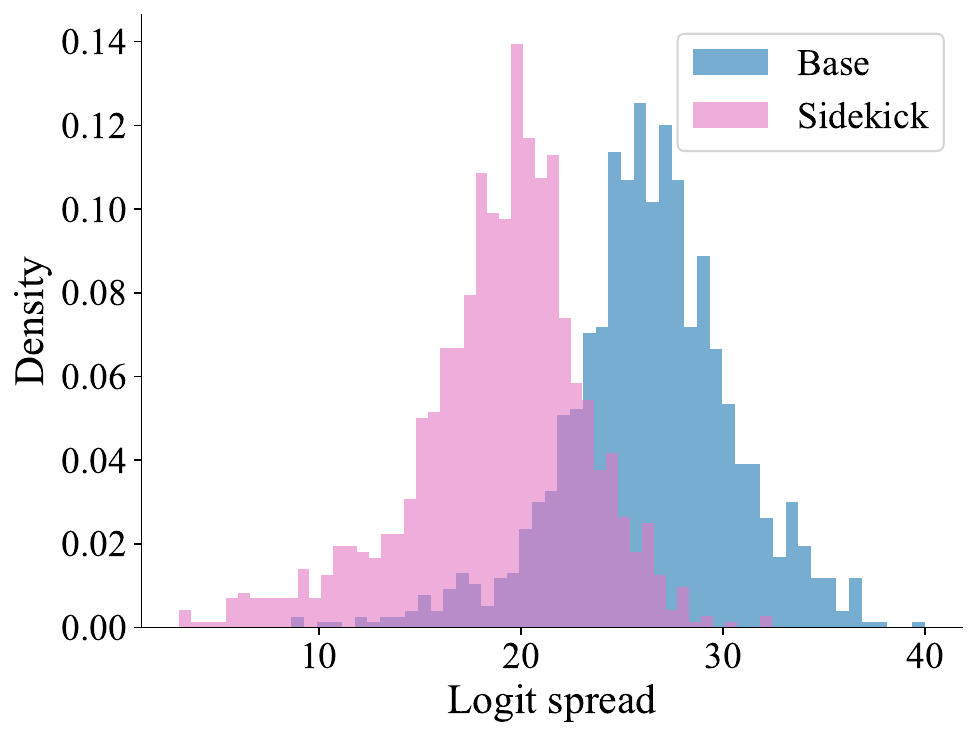}
    
    \caption{Logit (left) and logit spread (right) distributions of base and sidekick models on the CSQA dataset.}
    \label{fig:logits_distribution_base_sidekick}
\end{figure}

\newpage
\section{Supplementary Results}
\label{app:sup_results}

In this section, we provide the results of additional experiments to support our findings in the main text. Their interpretations are made in the respective sections of the paper.

\begin{table}[htbp]
\caption{Changes in test NLL produced by temperature scaling and the
additional changes produced by Duo. Negative values indicate reductions
in NLL and therefore improved performance.}
\label{tab:duo_vs_temperature_scaling}

\centering
\footnotesize
\setlength{\tabcolsep}{4pt}

\begin{tabular}{llcccccc}
\toprule
Comparison & Base method & ARC-C & ARC-E & CSQA & OBQA & RACE & SciQ \\
\midrule

\multirow{3}{*}{Base TS $-$ Base}
  & Plain LoRA
  & $-1.579$
  & $-0.442$
  & $-2.032$
  & $-0.634$
  & $-0.072$
  & $-0.150$ \\

  & Laplace-LoRA
  & $-0.725$
  & $-0.210$
  & $-0.941$
  & $-0.275$
  & $-0.051$
  & $-0.091$ \\

  & IB-EDL
  & $-0.003$
  & $-0.006$
  & $-0.005$
  & $-0.014$
  & $-0.005$
  & $-0.001$ \\

\midrule

\multirow{3}{*}{Duo $-$ Base TS}
  & Plain LoRA
  & $-0.027$
  & $-0.025$
  & $-0.052$
  & $-0.054$
  & $-0.006$
  & $-0.016$ \\

  & Laplace-LoRA
  & $-0.023$
  & $-0.020$
  & $-0.038$
  & $-0.040$
  & $-0.003$
  & $-0.009$ \\

  & IB-EDL
  & $-0.035$
  & $-0.030$
  & $-0.064$
  & $-0.044$
  & $-0.005$
  & $-0.022$ \\

\bottomrule
\end{tabular}%
\end{table}

\begin{table}[htbp]
\caption{Calibration performance of baseline and Duo methods fine-tuned on Qwen2
7B (``B'' for Base) and Qwen2 0.5B (``S'' for Sidekick). Accuracy and
ECE are reported as percentages. The best results among the shown
methods are highlighted in \textbf{bold}, and the second-best results
are \underline{underlined}. Additionally, $\pm$ denotes the standard
deviation across five runs with different seeds.}

\centering
\footnotesize
\setlength{\tabcolsep}{5pt}
\label{tab:additional_experiments_smaller_sidekick}

\resizebox{\textwidth}{!}{%
\begin{tabular}{llcccccc}
\toprule
Metric & Method & ARC-C & ARC-E & CSQA & OBQA & RACE & SciQ \\
\midrule

\multirow{4}{*}{NLL $\downarrow$}
  & B
  & $2.075_{\pm0.041}$
  & $0.655_{\pm0.065}$
  & $2.557_{\pm0.192}$
  & $0.991_{\pm0.080}$
  & $0.340_{\pm0.009}$
  & $0.333_{\pm0.018}$ \\

  & B TS
  & $\underline{0.496_{\pm0.011}}$
  & $\underline{0.213_{\pm0.015}}$
  & $\underline{0.525_{\pm0.017}}$
  & $\underline{0.357_{\pm0.012}}$
  & $\underline{0.268_{\pm0.005}}$
  & $\underline{0.183_{\pm0.008}}$ \\

  & \cellcolor{gray!15}Duo
  & \cellcolor{gray!15}$\mathbf{0.487_{\pm0.011}}$
  & \cellcolor{gray!15}$\mathbf{0.201_{\pm0.012}}$
  & \cellcolor{gray!15}$\mathbf{0.502_{\pm0.014}}$
  & \cellcolor{gray!15}$\mathbf{0.341_{\pm0.008}}$
  & \cellcolor{gray!15}$\mathbf{0.265_{\pm0.004}}$
  & \cellcolor{gray!15}$\mathbf{0.175_{\pm0.006}}$ \\

  & B DE
  & $1.682_{\pm0.069}$
  & $0.538_{\pm0.029}$
  & $1.925_{\pm0.101}$
  & $0.750_{\pm0.054}$
  & $0.319_{\pm0.006}$
  & $0.295_{\pm0.007}$ \\

\midrule

\multirow{4}{*}{Acc $\uparrow$}
  & B
  & $86.59_{\pm0.52}$
  & $95.08_{\pm0.40}$
  & $85.34_{\pm0.41}$
  & $\mathbf{90.24_{\pm0.17}}$
  & $\underline{91.89_{\pm0.11}}$
  & $94.96_{\pm0.21}$ \\

  & B TS
  & $86.59_{\pm0.52}$
  & $95.08_{\pm0.40}$
  & $85.34_{\pm0.41}$
  & $\mathbf{90.24_{\pm0.17}}$
  & $\underline{91.89_{\pm0.11}}$
  & $94.96_{\pm0.21}$ \\

  & \cellcolor{gray!15}Duo
  & \cellcolor{gray!15}$\underline{86.86_{\pm0.48}}$
  & \cellcolor{gray!15}$\mathbf{95.13_{\pm0.40}}$
  & \cellcolor{gray!15}$\underline{85.41_{\pm0.42}}$
  & \cellcolor{gray!15}$\underline{90.20_{\pm0.24}}$
  & \cellcolor{gray!15}$\underline{91.89_{\pm0.22}}$
  & \cellcolor{gray!15}$\underline{95.06_{\pm0.25}}$ \\

  & B DE
  & $\mathbf{87.15_{\pm0.43}}$
  & $\underline{95.09_{\pm0.14}}$
  & $\mathbf{85.81_{\pm0.35}}$
  & $\underline{90.20_{\pm0.32}}$
  & $\mathbf{91.94_{\pm0.23}}$
  & $\mathbf{95.12_{\pm0.25}}$ \\

\midrule

\multirow{4}{*}{ECE $\downarrow$}
  & B
  & $13.05_{\pm0.51}$
  & $4.74_{\pm0.42}$
  & $14.38_{\pm0.41}$
  & $9.55_{\pm0.32}$
  & $5.63_{\pm0.15}$
  & $4.75_{\pm0.13}$ \\

  & B TS
  & $\underline{4.35_{\pm0.84}}$
  & $\underline{1.53_{\pm0.21}}$
  & $\underline{3.54_{\pm0.63}}$
  & $\underline{4.13_{\pm1.07}}$
  & $\underline{2.73_{\pm0.30}}$
  & $\underline{3.07_{\pm0.53}}$ \\

  & \cellcolor{gray!15}Duo
  & \cellcolor{gray!15}$\mathbf{3.17_{\pm0.67}}$
  & \cellcolor{gray!15}$\mathbf{1.22_{\pm0.19}}$
  & \cellcolor{gray!15}$\mathbf{2.91_{\pm0.36}}$
  & \cellcolor{gray!15}$\mathbf{3.65_{\pm1.45}}$
  & \cellcolor{gray!15}$\mathbf{2.64_{\pm0.24}}$
  & \cellcolor{gray!15}$\mathbf{2.80_{\pm0.24}}$ \\

  & B DE
  & $10.11_{\pm0.93}$
  & $3.76_{\pm0.13}$
  & $10.35_{\pm0.57}$
  & $7.14_{\pm0.22}$
  & $5.17_{\pm0.25}$
  & $4.12_{\pm0.07}$ \\

\bottomrule
\end{tabular}%
}
\vspace{-0.8\baselineskip}
\end{table}

\begin{table}[htbp]
\caption{Calibration performance of baseline and Duo methods fine-tuned on Llama-3.1-8B (``B'' for Base) and Llama-3.2-1B (``S'' for Sidekick). Arrows
($\uparrow$ or $\downarrow$) indicate the direction of better
performance. The best results are highlighted in \textbf{bold}, and
the second-best results are \underline{underlined}. Additionally,
$\pm$ denotes the standard deviation over five runs with five
different seeds.}

\centering
\footnotesize
\setlength{\tabcolsep}{5pt}
\label{tab:additional_experiments_llama}

\resizebox{\textwidth}{!}{%
\begin{tabular}{llcccccc}
\toprule
Metric & Method & ARC-C & ARC-E & CSQA & OBQA & RACE & SciQ \\
\midrule

\multirow{5}{*}{NLL $\downarrow$}
  & B
  & $3.212_{\pm0.058}$
  & $1.132_{\pm0.079}$
  & $2.635_{\pm0.091}$
  & $1.326_{\pm0.080}$
  & $0.441_{\pm0.009}$
  & $0.395_{\pm0.032}$ \\

  & B TS
  & $\underline{0.721_{\pm0.013}}$
  & $\underline{0.330_{\pm0.016}}$
  & $\underline{0.650_{\pm0.008}}$
  & $\underline{0.466_{\pm0.020}}$
  & $\underline{0.339_{\pm0.004}}$
  & $\underline{0.204_{\pm0.009}}$ \\

  & \cellcolor{gray!15}Duo
  & \cellcolor{gray!15}$\mathbf{0.701_{\pm0.011}}$
  & \cellcolor{gray!15}$\mathbf{0.309_{\pm0.013}}$
  & \cellcolor{gray!15}$\mathbf{0.608_{\pm0.013}}$
  & \cellcolor{gray!15}$\mathbf{0.437_{\pm0.018}}$
  & \cellcolor{gray!15}$\mathbf{0.333_{\pm0.003}}$
  & \cellcolor{gray!15}$\mathbf{0.185_{\pm0.009}}$ \\

  & S
  & $6.631_{\pm0.381}$
  & $3.330_{\pm0.115}$
  & $2.912_{\pm0.161}$
  & $2.152_{\pm0.151}$
  & $0.835_{\pm0.016}$
  & $0.633_{\pm0.018}$ \\

  & B DE
  & $2.639_{\pm0.028}$
  & $0.876_{\pm0.067}$
  & $2.005_{\pm0.054}$
  & $0.991_{\pm0.042}$
  & $0.399_{\pm0.007}$
  & $0.351_{\pm0.016}$ \\

\midrule

\multirow{5}{*}{Acc $\uparrow$}
  & B
  & $80.27_{\pm0.71}$
  & $92.34_{\pm0.42}$
  & $82.28_{\pm0.36}$
  & $\underline{87.16_{\pm0.67}}$
  & $89.69_{\pm0.28}$
  & $\underline{94.44_{\pm0.17}}$ \\

  & B TS
  & $80.27_{\pm0.71}$
  & $92.34_{\pm0.42}$
  & $82.28_{\pm0.36}$
  & $\underline{87.16_{\pm0.67}}$
  & $89.69_{\pm0.28}$
  & $\underline{94.44_{\pm0.17}}$ \\

  & \cellcolor{gray!15}Duo
  & \cellcolor{gray!15}$\underline{80.44_{\pm0.66}}$
  & \cellcolor{gray!15}$\underline{92.37_{\pm0.38}}$
  & \cellcolor{gray!15}$\underline{82.38_{\pm0.59}}$
  & \cellcolor{gray!15}$86.84_{\pm0.82}$
  & \cellcolor{gray!15}$\underline{89.71_{\pm0.38}}$
  & \cellcolor{gray!15}$94.40_{\pm0.17}$ \\

  & S
  & $52.44_{\pm0.80}$
  & $74.82_{\pm0.66}$
  & $69.78_{\pm1.06}$
  & $69.68_{\pm1.51}$
  & $75.82_{\pm0.47}$
  & $87.58_{\pm0.58}$ \\

  & B DE
  & $\mathbf{80.85_{\pm0.48}}$
  & $\mathbf{92.95_{\pm0.40}}$
  & $\mathbf{82.80_{\pm0.34}}$
  & $\mathbf{87.20_{\pm0.32}}$
  & $\mathbf{90.10_{\pm0.10}}$
  & $\mathbf{94.54_{\pm0.32}}$ \\

\midrule

\multirow{5}{*}{ECE $\downarrow$}
  & B
  & $19.57_{\pm0.66}$
  & $7.52_{\pm0.46}$
  & $17.54_{\pm0.31}$
  & $12.86_{\pm0.73}$
  & $7.36_{\pm0.31}$
  & $5.19_{\pm0.25}$ \\

  & B TS
  & $\mathbf{5.39_{\pm1.06}}$
  & $\underline{2.41_{\pm0.47}}$
  & $\underline{4.64_{\pm1.25}}$
  & $\underline{4.87_{\pm1.09}}$
  & $\underline{4.55_{\pm0.33}}$
  & $\underline{3.32_{\pm0.68}}$ \\

  & \cellcolor{gray!15}Duo
  & \cellcolor{gray!15}$\underline{5.43_{\pm0.29}}$
  & \cellcolor{gray!15}$\mathbf{2.05_{\pm0.38}}$
  & \cellcolor{gray!15}$\mathbf{3.57_{\pm0.96}}$
  & \cellcolor{gray!15}$\mathbf{4.42_{\pm0.73}}$
  & \cellcolor{gray!15}$\mathbf{2.87_{\pm0.40}}$
  & \cellcolor{gray!15}$\mathbf{1.75_{\pm0.25}}$ \\

  & S
  & $46.44_{\pm0.85}$
  & $24.57_{\pm0.66}$
  & $28.70_{\pm1.02}$
  & $28.24_{\pm1.59}$
  & $14.03_{\pm0.47}$
  & $10.43_{\pm0.57}$ \\

  & B DE
  & $15.05_{\pm0.58}$
  & $5.40_{\pm0.26}$
  & $12.49_{\pm0.43}$
  & $8.93_{\pm0.24}$
  & $6.04_{\pm0.22}$
  & $4.64_{\pm0.21}$ \\

\bottomrule
\end{tabular}%
}
\end{table}

\newpage
\begin{table}[htbp]
\caption{Calibration performance of baseline and Duo methods fine-tuned on Llama-3.1-8B (``B'' for Base) and Qwen2-1.5B (``S'' for Sidekick). Arrows
($\uparrow$ or $\downarrow$) indicate the direction of better
performance. The best results are highlighted in \textbf{bold}, and
the second-best results are \underline{underlined}. Additionally,
$\pm$ denotes the standard deviation over five runs with five
different seeds.}

\centering
\footnotesize
\setlength{\tabcolsep}{5pt}
\label{tab:additional_experiments_cross_family}

\resizebox{\textwidth}{!}{%
\begin{tabular}{llcccccc}
\toprule
Metric & Method & ARC-C & ARC-E & CSQA & OBQA & RACE & SciQ \\
\midrule

\multirow{5}{*}{NLL $\downarrow$}
  & B
  & $3.212_{\pm0.058}$
  & $1.132_{\pm0.079}$
  & $2.635_{\pm0.091}$
  & $1.326_{\pm0.080}$
  & $0.441_{\pm0.009}$
  & $0.395_{\pm0.032}$ \\

  & B TS
  & $\underline{0.721_{\pm0.013}}$
  & $\underline{0.330_{\pm0.016}}$
  & $\underline{0.650_{\pm0.008}}$
  & $\underline{0.466_{\pm0.020}}$
  & $\underline{0.339_{\pm0.004}}$
  & $\underline{0.204_{\pm0.009}}$ \\

  & \cellcolor{gray!15}Duo
  & \cellcolor{gray!15}$\mathbf{0.633_{\pm0.011}}$
  & \cellcolor{gray!15}$\mathbf{0.264_{\pm0.010}}$
  & \cellcolor{gray!15}$\mathbf{0.576_{\pm0.006}}$
  & \cellcolor{gray!15}$\mathbf{0.385_{\pm0.014}}$
  & \cellcolor{gray!15}$\mathbf{0.314_{\pm0.004}}$
  & \cellcolor{gray!15}$\mathbf{0.179_{\pm0.006}}$ \\

  & S
  & $4.234_{\pm0.046}$
  & $1.968_{\pm0.100}$
  & $3.493_{\pm0.182}$
  & $1.497_{\pm0.076}$
  & $0.590_{\pm0.021}$
  & $0.546_{\pm0.022}$ \\

  & B DE
  & $2.639_{\pm0.028}$
  & $0.876_{\pm0.067}$
  & $2.005_{\pm0.054}$
  & $0.991_{\pm0.042}$
  & $0.399_{\pm0.007}$
  & $0.351_{\pm0.016}$ \\

\midrule

\multirow{5}{*}{Acc $\uparrow$}
  & B
  & $80.27_{\pm0.71}$
  & $92.34_{\pm0.42}$
  & $82.28_{\pm0.36}$
  & $87.16_{\pm0.67}$
  & $89.69_{\pm0.28}$
  & $94.44_{\pm0.17}$ \\

  & B TS
  & $80.27_{\pm0.71}$
  & $92.34_{\pm0.42}$
  & $82.28_{\pm0.36}$
  & $87.16_{\pm0.67}$
  & $89.69_{\pm0.28}$
  & $94.44_{\pm0.17}$ \\

  & \cellcolor{gray!15}Duo
  & \cellcolor{gray!15}$\underline{80.63_{\pm0.40}}$
  & \cellcolor{gray!15}$\underline{92.69_{\pm0.35}}$
  & \cellcolor{gray!15}$\underline{82.47_{\pm0.21}}$
  & \cellcolor{gray!15}$\mathbf{87.28_{\pm0.69}}$
  & \cellcolor{gray!15}$\underline{89.79_{\pm0.36}}$
  & \cellcolor{gray!15}$\mathbf{94.60_{\pm0.37}}$ \\

  & S
  & $71.54_{\pm0.90}$
  & $85.51_{\pm0.61}$
  & $74.63_{\pm0.27}$
  & $80.32_{\pm0.52}$
  & $84.29_{\pm0.42}$
  & $90.04_{\pm0.15}$ \\

  & B DE
  & $\mathbf{80.85_{\pm0.48}}$
  & $\mathbf{92.95_{\pm0.40}}$
  & $\mathbf{82.80_{\pm0.34}}$
  & $\underline{87.20_{\pm0.32}}$
  & $\mathbf{90.10_{\pm0.10}}$
  & $\underline{94.54_{\pm0.32}}$ \\

\midrule

\multirow{5}{*}{ECE $\downarrow$}
  & B
  & $19.57_{\pm0.66}$
  & $7.52_{\pm0.46}$
  & $17.54_{\pm0.31}$
  & $12.86_{\pm0.73}$
  & $7.36_{\pm0.31}$
  & $5.19_{\pm0.25}$ \\

  & B TS
  & $\underline{5.39_{\pm1.06}}$
  & $\underline{2.41_{\pm0.47}}$
  & $\underline{4.64_{\pm1.25}}$
  & $\underline{4.87_{\pm1.09}}$
  & $\underline{4.55_{\pm0.33}}$
  & $\underline{3.32_{\pm0.68}}$ \\

  & \cellcolor{gray!15}Duo
  & \cellcolor{gray!15}$\mathbf{5.27_{\pm0.99}}$
  & \cellcolor{gray!15}$\mathbf{1.28_{\pm0.16}}$
  & \cellcolor{gray!15}$\mathbf{3.00_{\pm0.54}}$
  & \cellcolor{gray!15}$\mathbf{4.15_{\pm0.81}}$
  & \cellcolor{gray!15}$\mathbf{3.19_{\pm0.30}}$
  & \cellcolor{gray!15}$\mathbf{2.11_{\pm0.33}}$ \\

  & S
  & $27.73_{\pm0.96}$
  & $14.16_{\pm0.59}$
  & $24.61_{\pm0.36}$
  & $18.57_{\pm0.65}$
  & $10.11_{\pm0.47}$
  & $8.69_{\pm0.14}$ \\

  & B DE
  & $15.05_{\pm0.58}$
  & $5.40_{\pm0.26}$
  & $12.49_{\pm0.43}$
  & $8.93_{\pm0.24}$
  & $6.04_{\pm0.22}$
  & $4.64_{\pm0.21}$ \\

\bottomrule
\end{tabular}%
}
\end{table}

\begin{table}[htbp]
\caption{Learned Duo coefficients at different validation subset sizes.
The full-data results are highlighted in gray as a reference.}
\label{tab:coefficient_sensitivity}

\centering
\footnotesize
\setlength{\tabcolsep}{4.5pt}

\resizebox{\textwidth}{!}{%
\begin{tabular}{llcccccc}
\toprule
Coefficient & Val. subset & ARC-C & ARC-E & CSQA & OBQA & RACE & SciQ \\
\midrule

\multirow{3}{*}{$\alpha$}
  & 20\%
  & $0.175_{\pm0.043}$
  & $0.230_{\pm0.039}$
  & $0.122_{\pm0.014}$
  & $0.228_{\pm0.042}$
  & $0.582_{\pm0.069}$
  & $0.343_{\pm0.053}$ \\

  & 40\%
  & $0.162_{\pm0.020}$
  & $0.224_{\pm0.029}$
  & $0.119_{\pm0.010}$
  & $0.225_{\pm0.031}$
  & $0.572_{\pm0.047}$
  & $0.327_{\pm0.023}$ \\

  & \cellcolor{gray!15}100\%
  & \cellcolor{gray!15}$0.162_{\pm0.006}$
  & \cellcolor{gray!15}$0.220_{\pm0.012}$
  & \cellcolor{gray!15}$0.120_{\pm0.008}$
  & \cellcolor{gray!15}$0.219_{\pm0.022}$
  & \cellcolor{gray!15}$0.568_{\pm0.022}$
  & \cellcolor{gray!15}$0.330_{\pm0.013}$ \\

\midrule

\multirow{3}{*}{$\beta$}
  & 20\%
  & $0.055_{\pm0.039}$
  & $0.064_{\pm0.031}$
  & $0.070_{\pm0.015}$
  & $0.118_{\pm0.050}$
  & $0.080_{\pm0.068}$
  & $0.126_{\pm0.067}$ \\

  & 40\%
  & $0.043_{\pm0.020}$
  & $0.071_{\pm0.027}$
  & $0.072_{\pm0.011}$
  & $0.127_{\pm0.032}$
  & $0.091_{\pm0.042}$
  & $0.135_{\pm0.033}$ \\

  & \cellcolor{gray!15}100\%
  & \cellcolor{gray!15}$0.047_{\pm0.010}$
  & \cellcolor{gray!15}$0.072_{\pm0.012}$
  & \cellcolor{gray!15}$0.071_{\pm0.005}$
  & \cellcolor{gray!15}$0.126_{\pm0.017}$
  & \cellcolor{gray!15}$0.083_{\pm0.017}$
  & \cellcolor{gray!15}$0.135_{\pm0.020}$ \\

\bottomrule
\end{tabular}%
}
\end{table}

\begin{table}[htbp]
\caption{Test NLL obtained by fitting Duo on different validation subset sizes. The full-data results are highlighted
in gray.}
\label{tab:nll_sensitivity}

\centering
\footnotesize
\setlength{\tabcolsep}{5pt}

\begin{tabular}{lcccccc}
\toprule
Val. subset & ARC-C & ARC-E & CSQA & OBQA & RACE & SciQ \\
\midrule

20\%
  & $0.498_{\pm0.067}$
  & $0.196_{\pm0.015}$
  & $0.477_{\pm0.014}$
  & $0.313_{\pm0.014}$
  & $0.266_{\pm0.007}$
  & $0.171_{\pm0.008}$ \\

40\%
  & ${0.475_{\pm0.016}}$
  & ${0.192_{\pm0.015}}$
  & ${0.475_{\pm0.013}}$
  & ${0.308_{\pm0.012}}$
  & ${0.263_{\pm0.006}}$
  & ${0.168_{\pm0.006}}$ \\

\cellcolor{gray!15}100\%
  & \cellcolor{gray!15}${0.469_{\pm0.009}}$
  & \cellcolor{gray!15}${0.188_{\pm0.014}}$
  & \cellcolor{gray!15}${0.473_{\pm0.014}}$
  & \cellcolor{gray!15}${0.302_{\pm0.008}}$
  & \cellcolor{gray!15}${0.262_{\pm0.006}}$
  & \cellcolor{gray!15}${0.167_{\pm0.006}}$ \\

\bottomrule
\end{tabular}%
\end{table}

\newpage
\begin{table}[htbp]
\caption{Test-set fractions and contributions to the total NLL
improvement of Duo over Base TS.}
\label{tab:prediction_change_ablation}

\centering
\footnotesize
\setlength{\tabcolsep}{4pt}

\begin{tabular}{llcccccc}
\toprule
Measure & Subset & ARC-C & ARC-E & CSQA & OBQA & RACE & SciQ \\
\midrule

\multirow{4}{*}{Test-set fraction}
  & Prediction unchanged
  & $0.9915$
  & $0.9950$
  & $0.9789$
  & $0.9896$
  & $0.9943$
  & $0.9934$ \\

  & Prediction changed
  & $0.0085$
  & $0.0050$
  & $0.0211$
  & $0.0104$
  & $0.0057$
  & $0.0066$ \\

  & Changed: wrong to right
  & $0.0044$
  & $0.0024$
  & $0.0118$
  & $0.0052$
  & $0.0019$
  & $0.0046$ \\

  & Changed: right to wrong
  & $0.0022$
  & $0.0018$
  & $0.0079$
  & $0.0036$
  & $0.0026$
  & $0.0018$ \\

\midrule

\multirow{2}{*}{Share of NLL gain}
  & Prediction unchanged
  & $0.9704$
  & $0.9781$
  & $0.9257$
  & $0.9645$
  & $1.0360$
  & $0.8909$ \\

  & Prediction changed
  & $0.0296$
  & $0.0219$
  & $0.0743$
  & $0.0355$
  & $-0.0360$
  & $0.1091$ \\

\bottomrule
\end{tabular}%
\end{table}

\begin{table}[htbp]
\caption{Variants used to separate global scaling from
asymmetric weighting.}
\label{tab:scaling_weighting_variants}

\centering
\footnotesize
\setlength{\tabcolsep}{7pt}
\renewcommand{\arraystretch}{1.1}

\begin{tabular}{lcl}
\toprule
Variant & Parameters & Fusion rule \\
\midrule

Base
  & 0
  & $\mathbf{z}_b$ \\

Base TS
  & 1
  & $\alpha\mathbf{z}_b$ \\

Symmetric learned
  & 1
  & $\gamma(\mathbf{z}_b+\mathbf{z}_s)$ \\

\cellcolor{gray!15}Duo
  & \cellcolor{gray!15}2
  & \cellcolor{gray!15}
    $\alpha\mathbf{z}_b+\beta\mathbf{z}_s$ \\

Duo oracle
  & 2
  & $\alpha\mathbf{z}_b+\beta\mathbf{z}_s$,
    fitted on test \\

\bottomrule
\end{tabular}
\end{table}

\begin{table}[htbp]
\caption{Test NLL for the ablation separating global scaling from
asymmetric weighting. Bold and underlined values indicate the best and
second-best validation-fitted methods, respectively. Duo oracle is
fitted on the test set and shown only as a lower-bound reference.
The symbol $\pm$ denotes the standard deviation over five runs.}
\label{tab:scaling_weighting_nll}

\centering
\footnotesize
\setlength{\tabcolsep}{5pt}

\resizebox{\textwidth}{!}{%
\begin{tabular}{lcccccc}
\toprule
Variant & ARC-C & ARC-E & CSQA & OBQA & RACE & SciQ \\
\midrule

Base
  & $2.075_{\pm0.041}$
  & $0.655_{\pm0.065}$
  & $2.557_{\pm0.192}$
  & $0.990_{\pm0.080}$
  & $0.340_{\pm0.009}$
  & $0.333_{\pm0.018}$ \\

Base TS
  & $\underline{0.496_{\pm0.010}}$
  & $\underline{0.213_{\pm0.015}}$
  & $0.525_{\pm0.017}$
  & $0.357_{\pm0.012}$
  & $\underline{0.268_{\pm0.005}}$
  & $0.183_{\pm0.008}$ \\

Symmetric learned
  & $0.513_{\pm0.011}$
  & $0.216_{\pm0.012}$
  & $\underline{0.486_{\pm0.010}}$
  & $\underline{0.309_{\pm0.008}}$
  & $0.281_{\pm0.006}$
  & $\underline{0.173_{\pm0.005}}$ \\

\cellcolor{gray!15}Duo
  & \cellcolor{gray!15}$\mathbf{0.469_{\pm0.009}}$
  & \cellcolor{gray!15}$\mathbf{0.188_{\pm0.014}}$
  & \cellcolor{gray!15}$\mathbf{0.473_{\pm0.014}}$
  & \cellcolor{gray!15}$\mathbf{0.302_{\pm0.007}}$
  & \cellcolor{gray!15}$\mathbf{0.262_{\pm0.006}}$
  & \cellcolor{gray!15}$\mathbf{0.167_{\pm0.006}}$ \\

Duo oracle (lower bound)
  & $0.465_{\pm0.007}$
  & $0.186_{\pm0.013}$
  & $0.473_{\pm0.013}$
  & $0.300_{\pm0.007}$
  & $0.260_{\pm0.004}$
  & $0.165_{\pm0.006}$ \\

\bottomrule
\end{tabular}%
}
\end{table}

\newpage

\FloatBarrier
\section{Full numerical results}
All tables in this appendix report the main configuration (Qwen2-7B Base, Qwen2-1.5B Sidekick); results for the other model pairings are in Appendix~\ref{app:sup_results}.
\subsection{Compatibility with calibration-aware methods}
\label{app:compat}
Table~\ref{tab:compat} reports the per-(metric, dataset) numbers that underlie Figure~\ref{fig:duo_compatibility} of the main text. Within each B $X$ vs Duo $X$ pair, Duo $X$ achieves the better NLL on every dataset across all three calibration families. On ECE, Duo $X$ improves over B $X$ in 17 of 18 cells, with the single exception of CSQA-BLoB where the two methods are essentially tied (B BLoB $2.04$ vs Duo BLoB $2.06$, within standard deviation). Accuracy is preserved or improved in 16 of 18 cells, with two minor exceptions where Base wins by a small margin (OBQA-BLoB and RACE-LA).
\begin{table}[t]
\caption{Compatibility of the Duo procedure with calibration-aware baseline methods. Each block compares the calibration-aware base predictor B $X$ against its Duo counterpart Duo $X$ for $X\in\{\text{LA},\text{BLoB},\text{IB-EDL}\}$. Within each pair, the better entry per (metric, dataset) is in \textbf{bold}; ties are bolded on both rows. All metrics are mean $\pm$ SD over 5 repetitions. Plain-LoRA Duo and a Duo built from temperature-scaled Base and Sidekick are omitted because both reduce to the standard Duo of Table~\ref{tab:results_overall} (see remark in Section~\ref{sec:compat}).}
\label{tab:compat}
\centering
\footnotesize
\setlength{\tabcolsep}{5pt}
\resizebox{\textwidth}{!}{
\begin{tabular}{llcccccc}
\toprule
Metric & Method & ARC-C & ARC-E & CSQA & OBQA & RACE & SciQ \\
\midrule
\multirow{6}{*}{NLL $\downarrow$}
  & B LA       & $1.212_{\pm0.052}$ & $0.401_{\pm0.029}$ & $1.419_{\pm0.096}$ & $0.602_{\pm0.068}$ & $0.308_{\pm0.006}$ & $0.259_{\pm0.016}$ \\
  & Duo LA     & $\mathbf{0.438}_{\pm0.006}$ & $\mathbf{0.168}_{\pm0.009}$ & $\mathbf{0.432}_{\pm0.010}$ & $\mathbf{0.276}_{\pm0.019}$ & $\mathbf{0.255}_{\pm0.005}$ & $\mathbf{0.158}_{\pm0.006}$ \\
\cmidrule(l){2-8}
  & B BLoB     & $1.347_{\pm0.101}$ & $0.183_{\pm0.039}$ & $0.361_{\pm0.008}$ & $0.240_{\pm0.016}$ & $0.247_{\pm0.009}$ & $0.136_{\pm0.002}$ \\
  & Duo BLoB   & $\mathbf{0.443}_{\pm0.014}$ & $\mathbf{0.152}_{\pm0.012}$ & $\mathbf{0.355}_{\pm0.007}$ & $\mathbf{0.234}_{\pm0.017}$ & $\mathbf{0.245}_{\pm0.009}$ & $\mathbf{0.135}_{\pm0.003}$ \\
\cmidrule(l){2-8}
  & B IB-EDL   & $0.547_{\pm0.018}$ & $0.220_{\pm0.021}$ & $0.556_{\pm0.017}$ & $0.368_{\pm0.016}$ & $0.267_{\pm0.003}$ & $0.205_{\pm0.005}$ \\
  & Duo IB-EDL & $\mathbf{0.509}_{\pm0.015}$ & $\mathbf{0.183}_{\pm0.018}$ & $\mathbf{0.487}_{\pm0.018}$ & $\mathbf{0.310}_{\pm0.016}$ & $\mathbf{0.257}_{\pm0.003}$ & $\mathbf{0.182}_{\pm0.005}$ \\
\midrule
\multirow{6}{*}{Acc $\uparrow$}
  & B LA       & $86.66_{\pm0.48}$ & $95.11_{\pm0.32}$ & $85.32_{\pm0.42}$ & $90.04_{\pm0.14}$ & $\mathbf{91.80}_{\pm0.15}$ & $94.84_{\pm0.23}$ \\
  & Duo LA     & $\mathbf{87.12}_{\pm0.32}$ & $\mathbf{95.23}_{\pm0.40}$ & $\mathbf{85.81}_{\pm0.67}$ & $\mathbf{90.64}_{\pm0.17}$ & $91.75_{\pm0.49}$ & $\mathbf{95.22}_{\pm0.33}$ \\
\cmidrule(l){2-8}
  & B BLoB     & $86.31_{\pm0.67}$ & $94.89_{\pm0.38}$ & $86.55_{\pm0.61}$ & $\mathbf{91.84}_{\pm0.83}$ & $\mathbf{91.57}_{\pm0.59}$ & $95.42_{\pm0.08}$ \\
  & Duo BLoB   & $\mathbf{86.55}_{\pm0.30}$ & $\mathbf{94.97}_{\pm0.50}$ & $\mathbf{86.58}_{\pm0.62}$ & $91.36_{\pm0.83}$ & $\mathbf{91.57}_{\pm0.59}$ & $\mathbf{95.44}_{\pm0.11}$ \\
\cmidrule(l){2-8}
  & B IB-EDL   & $86.28_{\pm0.60}$ & $95.23_{\pm0.17}$ & $85.39_{\pm0.52}$ & $90.84_{\pm0.38}$ & $91.85_{\pm0.26}$ & $95.02_{\pm0.22}$ \\
  & Duo IB-EDL & $\mathbf{86.55}_{\pm0.67}$ & $\mathbf{95.45}_{\pm0.24}$ & $\mathbf{85.57}_{\pm0.76}$ & $\mathbf{90.92}_{\pm0.59}$ & $\mathbf{91.88}_{\pm0.26}$ & $\mathbf{95.24}_{\pm0.32}$ \\
\midrule
\multirow{6}{*}{ECE $\downarrow$}
  & B LA       & $9.79_{\pm0.39}$  & $3.55_{\pm0.31}$  & $10.95_{\pm0.72}$ & $7.03_{\pm0.42}$  & $4.77_{\pm0.23}$  & $3.81_{\pm0.25}$ \\
  & Duo LA     & $\mathbf{2.02}_{\pm0.41}$ & $\mathbf{1.08}_{\pm0.43}$ & $\mathbf{2.37}_{\pm0.76}$ & $\mathbf{2.84}_{\pm0.37}$ & $\mathbf{2.12}_{\pm0.25}$ & $\mathbf{1.71}_{\pm0.23}$ \\
\cmidrule(l){2-8}
  & B BLoB     & $12.27_{\pm0.67}$ & $5.48_{\pm2.90}$  & $\mathbf{2.04}_{\pm0.15}$ & $2.71_{\pm0.75}$  & $2.17_{\pm0.31}$  & $1.55_{\pm0.38}$ \\
  & Duo BLoB   & $\mathbf{4.24}_{\pm0.47}$ & $\mathbf{0.97}_{\pm0.21}$ & $2.06_{\pm0.30}$ & $\mathbf{2.64}_{\pm0.52}$ & $\mathbf{1.48}_{\pm0.35}$ & $\mathbf{1.39}_{\pm0.24}$ \\
\cmidrule(l){2-8}
  & B IB-EDL   & $4.88_{\pm1.04}$  & $3.90_{\pm1.01}$  & $5.61_{\pm0.78}$  & $9.37_{\pm1.20}$  & $3.84_{\pm0.55}$  & $2.89_{\pm0.65}$ \\
  & Duo IB-EDL & $\mathbf{2.73}_{\pm1.03}$ & $\mathbf{1.07}_{\pm0.14}$ & $\mathbf{3.65}_{\pm0.52}$ & $\mathbf{2.81}_{\pm0.68}$ & $\mathbf{2.10}_{\pm0.64}$ & $\mathbf{2.19}_{\pm0.41}$ \\
\bottomrule
\end{tabular}}
\end{table}

\subsection{Learned Duo fusion weights}
\label{app:duo_alpha_beta}
Table~\ref{tab:duo_alpha_beta} reports the fusion weights $(\alpha,\beta)$ for every Duo variant in this paper, aggregated as mean $\pm$ standard deviation over 5 repetitions. Two regimes emerge clearly. For plain Duo and Duo LA the optimizer settles on $\alpha,\beta\ll 1$, consistent with both base and sidekick logits being sharply overconfident: the optimizer compresses them in addition to combining them, in line with the trivial-Duo ablation of Section~\ref{sec:ablation}. For Duo BLoB and Duo IB-EDL the calibration-aware base has already softened the answer-token logits, so the optimizer leaves them near unit scale ($\alpha\approx 1$) and tunes $\beta$ to decide how much to incorporate the sidekick. On RACE and SciQ the BLoB sidekick is so saturated that $\beta$ collapses to the optimizer's lower bound ($10^{-6}$), and Duo BLoB reduces to B BLoB on those datasets.
\begin{table}[t]
\caption{Learned Duo fusion weights $(\alpha,\beta)$ on the validation logits of base and sidekick, mean $\pm$ SD over 5 repetitions. Values reported as $0.000$ for $\beta$ are at the optimizer's lower bound ($10^{-6}$).}
\label{tab:duo_alpha_beta}
\centering
\footnotesize
\setlength{\tabcolsep}{4pt}
\resizebox{\textwidth}{!}{
\begin{tabular}{llcccccc}
\toprule
Method & Weight & ARC-C & ARC-E & CSQA & OBQA & RACE & SciQ \\
\midrule
\multirow{2}{*}{Duo}
  & $\alpha$ & $0.162_{\pm0.006}$ & $0.220_{\pm0.012}$ & $0.120_{\pm0.008}$ & $0.219_{\pm0.022}$ & $0.568_{\pm0.022}$ & $0.330_{\pm0.013}$ \\
  & $\beta$  & $0.046_{\pm0.010}$ & $0.072_{\pm0.012}$ & $0.071_{\pm0.005}$ & $0.126_{\pm0.017}$ & $0.083_{\pm0.017}$ & $0.135_{\pm0.020}$ \\
\midrule
\multirow{2}{*}{Duo LA}
  & $\alpha$ & $0.227_{\pm0.012}$ & $0.289_{\pm0.015}$ & $0.167_{\pm0.008}$ & $0.281_{\pm0.015}$ & $0.626_{\pm0.027}$ & $0.399_{\pm0.014}$ \\
  & $\beta$  & $0.068_{\pm0.012}$ & $0.080_{\pm0.015}$ & $0.100_{\pm0.005}$ & $0.155_{\pm0.029}$ & $0.056_{\pm0.022}$ & $0.111_{\pm0.023}$ \\
\midrule
\multirow{2}{*}{Duo BLoB}
  & $\alpha$ & $0.228_{\pm0.013}$ & $1.540_{\pm0.118}$ & $0.922_{\pm0.041}$ & $1.022_{\pm0.097}$ & $1.116_{\pm0.010}$ & $1.031_{\pm0.032}$ \\
  & $\beta$  & $0.049_{\pm0.014}$ & $0.142_{\pm0.099}$ & $0.261_{\pm0.028}$ & $0.145_{\pm0.057}$ & $0.000_{\pm0.000}$ & $0.001_{\pm0.003}$ \\
\midrule
\multirow{2}{*}{Duo IB-EDL}
  & $\alpha$ & $0.960_{\pm0.053}$ & $0.936_{\pm0.095}$ & $0.865_{\pm0.030}$ & $0.981_{\pm0.045}$ & $1.062_{\pm0.022}$ & $0.820_{\pm0.023}$ \\
  & $\beta$  & $0.411_{\pm0.054}$ & $0.430_{\pm0.081}$ & $0.484_{\pm0.023}$ & $0.428_{\pm0.066}$ & $0.135_{\pm0.023}$ & $0.320_{\pm0.022}$ \\
\bottomrule
\end{tabular}}
\end{table}

\subsection{Mechanistic logit-space analysis of Asymmetric Duos}
\label{app:duo_improvement_points}
Table~\ref{tab:base_correct_wrong_all} reports the Base-Correct vs Base-Wrong decomposition of Section~\ref{sec:logit_distribution} for the remaining five benchmarks. The same asymmetry holds for every dataset: Duo trades a small confidence cost on Base's correct subset for a substantial reduction on Base's wrong subset.
\begin{table}[t]
\caption{NLL and ECE of Base and Duo, computed separately on test set samples of benchmark datasets that Base classifies correctly/incorrectly. Bold indicates the better value between Base and Duo within each subset.}
\centering
\footnotesize
\setlength{\tabcolsep}{10pt}
\label{tab:base_correct_wrong_all}
\sisetup{detect-weight=true, detect-family=true}
\begin{tabular}{lllS[table-format=2.3]S[table-format=2.3]}
\toprule
Metric & Dataset & Method & {Base Correct} & {Base Wrong}\\
\midrule
\multirow{10}{*}{NLL $\downarrow$}
& \multirow{2}{*}{ARC-C} & Base & \bfseries 0.001 & 16.426\\
&                        & Duo  & 0.090           & \bfseries 2.996\\
\cmidrule(lr){2-5}
& \multirow{2}{*}{ARC-E} & Base & \bfseries 0.001 & 13.313\\
&                        & Duo  & 0.041           & \bfseries 2.935\\
\cmidrule(lr){2-5}
& \multirow{2}{*}{OBQA}  & Base & \bfseries 0.001 & 11.055\\
&                        & Duo  & 0.087           & \bfseries 2.224\\
\cmidrule(lr){2-5}
& \multirow{2}{*}{RACE}  & Base & \bfseries 0.016 & 4.041\\
&                        & Duo  & 0.055           & \bfseries 2.638\\
\cmidrule(lr){2-5}
& \multirow{2}{*}{SciQ}  & Base & \bfseries 0.003 & 6.625\\
&                        & Duo  & 0.037           & \bfseries 2.702\\
\midrule
\multirow{10}{*}{ECE $\downarrow$}
& \multirow{2}{*}{ARC-C} & Base & \bfseries 0.08  & 97.53\\
&                        & Duo  & 8.13            & \bfseries 78.04\\
\cmidrule(lr){2-5}
& \multirow{2}{*}{ARC-E} & Base & \bfseries 0.06  & 98.40\\
&                        & Duo  & 3.47            & \bfseries 76.51\\
\cmidrule(lr){2-5}
& \multirow{2}{*}{OBQA}  & Base & \bfseries 0.11  & 99.50\\
&                        & Duo  & 6.88            & \bfseries 61.39\\
\cmidrule(lr){2-5}
& \multirow{2}{*}{RACE}  & Base & \bfseries 1.30  & 82.71\\
&                        & Duo  & 4.66            & \bfseries 72.57\\
\cmidrule(lr){2-5}
& \multirow{2}{*}{SciQ}  & Base & \bfseries 0.25  & 95.65\\
&                        & Duo  & 3.45            & \bfseries 74.08\\
\bottomrule
\end{tabular}
\end{table}

\section{Detailed implementation}
\label{app:det_imp}

A reproducible codebase for all experiments of this work is provided at \href{https://github.com/isarustu1-dotcom/asymmetric-llm-duos}{https://github.com/isarustu1-dotcom/asymmetric-llm-duos}.

\paragraph{Models.} We fine-tune Qwen2-7B, Qwen2-1.5B, and Qwen2-0.5B~\citep{qwen2} and, from a second model family, Llama-3.1-8B and Llama-3.2-1B~\citep{llama3}, all with LoRA~\citep{hu2022lora} via PEFT~\citep{peft} and Transformers~\citep{transformers}. All models are loaded in \texttt{bfloat16}. Neither family provides a dedicated padding token, so we reuse the end-of-sequence token for padding (\texttt{<|endoftext|>} for Qwen2, \texttt{<|end\_of\_text|>} for Llama~3).

\paragraph{Datasets.} In the experiments, six MCQA datasets are used, which are ARC-C and ARC-E~\citep{arc}, OpenBookQA (OBQA)~\citep{obqa}, CommonsenseQA (CSQA)~\citep{csqa}, SciQ~\citep{sciq}, and RACE~\citep{race}. These datasets are chosen to align with experimental settings in prior work, facilitating consistent and comparable evaluation across different methods~\mbox{\citep{wang2024blob, hu2022lora, li2025calibrating_llm}}. The target space is defined as tokens corresponding to different pre-determined options like A, B, C, D, and E.

\paragraph{Baselines.} Our method is compared to six different baselines: LoRA fine-tuned MAP predictions (\textbf{B}), a deep ensemble with two members (\textbf{B DE})~\citep{lakshminarayanan2017deep_ensembles, fort2019deepens}, Temperature Scaling (\textbf{B TS})~\citep{guo2017calibration}, Laplace-LoRA (\textbf{B LA})~\citep{yang2024bayesian}, IB-EDL (\textbf{B IB-EDL})~\citep{li2025calibrating_llm}, and BLoB (\textbf{B BLoB})~\citep{wang2024blob}. In order to test whether applying the Duo method to existing uncertainty-aware approaches can further lead to a performance boost, we additionally applied the Duo procedure to the outputs of the following baselines: Laplace-LoRA (\textbf{Duo LA}), IB-EDL (\textbf{Duo IB-EDL}), and BLoB (\textbf{Duo BLoB}).

\paragraph{Measuring the computational cost.} We use FLOPs to measure inference cost, as they provide a hardware-independent metric for comparing computational complexity across different methods. FLOPs are estimated from multiply-accumulate operations (MACs), where one MAC corresponds to two FLOPs. For the Duo framework, FLOPs are computed as the sum of the base and sidekick model inference costs. This allows for a fair comparison against baselines such as single models and deep ensembles of varying sizes. All estimates are obtained from standard model-level complexity calculations based on the model architecture and input sequence length.

\paragraph{LoRA hyperparameters.} We adapt both models with LoRA~\citep{hu2022lora} via the PEFT library~\citep{peft} and Transformers~\citep{transformers}. LoRA adapters are applied to the four attention projections of every transformer block, \texttt{q\_proj}, \texttt{k\_proj}, \texttt{v\_proj}, and \texttt{o\_proj}. We use rank $r=8$, scaling $\alpha_{\text{LoRA}}=16$, dropout $0.1$, and \texttt{bias = "lora\_only"}, with \texttt{task\_type = CAUSAL\_LM}. The same configuration is shared across all six datasets and all five models (Qwen2-0.5B/1.5B/7B, Llama-3.2-1B, Llama-3.1-8B).

\paragraph{Training details.} For all datasets, we use the original train/validation/test splits provided by their respective authors, accessed via the Hugging Face datasets library~\citep{transformers}; the sole exception
is CSQA, which lacks a dedicated validation split, so we reserve $10\%$
of its training data as the validation set. All MAP and IB-EDL runs use AdamW (HuggingFace \texttt{Trainer} defaults) with learning rate $5\times 10^{-5}$, a cosine schedule with $20$ warm-up steps, and \texttt{bfloat16} mixed precision. Per-device train and eval batch sizes are both $4$. We train for $10\,080$ optimizer steps on every dataset except CSQA, which uses $30\,000$ steps to compensate for its smaller training split. The maximum tokenized prompt length is $300$ tokens for ARC-C, ARC-E, CSQA, OBQA, and SciQ, and $512$ tokens for RACE (whose passages are longer). For MAP we use weight decay $0$; for IB-EDL we follow common evidential-training practice and add weight decay $10^{-3}$ together with gradient clipping at maximum global norm $20.0$. We adopt the MCQA prompt templates and the next-token answer-classification setup of the IB-EDL reference implementation~\citep{li2025calibrating_llm} unchanged, so that differences between methods are attributable to the calibration procedure rather than to prompt design. Each method is run with $5$ seeds and we report mean $\pm$ standard deviation across seeds. Validation and test answer-token logits at the final input position are saved once per run and reused by every post-hoc method below.

\paragraph{Computational details.}
All experiments were run on a DGX A100 cluster, using a single 40\,GB GPU per run; we did not use multi-GPU or distributed training. Fine-tuning a base model on one dataset and evaluating on its test set takes roughly 30--40 minutes, with the exception of CSQA, which uses a larger number of training steps and takes around 90 minutes. For each dataset, we trained 5 base models and 5 sidekick models with different random seeds and report the mean and standard deviation across these runs. Computing the Duo fusion weights on top of the fine-tuned models takes only seconds.

\paragraph{Asymmetric Duo fusion.} The fusion scales $(\alpha,\beta)$ are estimated on the validation logits with L-BFGS-B, initialized at $(\alpha,\beta) = (1.0, 1.0)$ and bounded below by $10^{-6}$ on each coordinate (no upper bound, no sum constraint). The loss is the validation NLL of $\mathrm{softmax}(\alpha z_{\text{base}} + \beta z_{\text{side}})$, evaluated in float64 with the standard log-sum-exp stabilization. The same optimizer settings are used for every Duo variant in the paper, including the calibration-aware Duos (Duo LA, Duo BLoB, Duo IB-EDL), where $\alpha$ and $\beta$ are fit on the validation logits of the corresponding calibration-aware Base and Sidekick. Logits from the two models are matched by example index before fitting.

\paragraph{Temperature scaling.} A single scalar temperature $T>0$ per model is fit on the validation NLL via L-BFGS-B in log-space (initialized at $T=1$) with the analytical gradient $\partial\,\mathrm{NLL}/\partial\log T = (1/T)\,\mathbb{E}_i[\,\ell_{i,y_i} - \mathbb{E}_{p_i}[\ell_i]\,]$, then applied to the test logits~\citep{guo2017calibration}. We optimize on float64 logits to avoid the gradient collapsing to zero under float32 precision.

\paragraph{Laplace-LoRA.} We use the official \texttt{bayesian-lora} package~\citep{yang2024bayesian}. Kronecker factors are computed on the training loader with at most $n_{\text{KFAC}}=10$ low-rank components per factor and a low-rank threshold of $100$. The prior variance $\sigma^2$ is initialized at $0.1$ and tuned by maximizing the model evidence~\citep{yang2024bayesian}. Predictive variances are produced via the Jacobian-mean approximation. We did not perform additional hyperparameter tuning beyond the package defaults.

\paragraph{IB-EDL implementation details.} We follow the IB-EDL formulation and implementation of \citet{li2025calibrating_llm}. The pre-evidence head is the language-model head of the LLM evaluated at the last input position and restricted to the answer-token columns; positivity is enforced with a softplus and a Dirichlet over the $K$ answer choices is constructed by adding $1$ to the evidence vector. The variational information bottleneck draws $K_{\text{IB}}=20$ Gaussian noise samples per training example, keeps $\sigma_{\text{mult}}=0$ at evaluation (i.e.\ no post-hoc evidence shrinkage), and uses the sum-of-squares Bayes-risk loss without label smoothing. The KL regularization weight is ramped up linearly starting at $10\%$ of training; we set the final regularization weight to $0$ since the IB term itself already regularizes the evidence. The information-bottleneck coefficient $\beta$ is the only IB-EDL hyperparameter we tune—per (model, dataset) pair—using the small grid in Table~\ref{tab:ibedl_beta} in Appendix~\ref{app:ibedl_beta}.

\paragraph{BLoB.} BLoB~\citep{wang2024blob} is run with the authors' official \texttt{bayesian-peft} codebase using its default settings, including $10$ Monte-Carlo samples at evaluation. We average the per-sample probabilities to form the Bayesian predictive and convert back to log-probabilities so that the saved BLoB logits live in the same log-probability space as the other methods.

\paragraph{Deep ensemble baseline (B DE).} For each dataset we form $5$ random ensembles of size $2$ from the $5$ available training seeds. Predictions are averaged in probability space, and metrics are then computed on each sampled pair and averaged over pairs.

\paragraph{Calibration metrics.} ECE is computed with $15$ equal-width confidence bins. NLL is the mean $-\log p_y$ over the test set, with a $10^{-12}$ floor on probabilities for numerical stability.

\subsection{Selection of the IB-EDL $\beta$}
\label{app:ibedl_beta}
The IB-EDL objective is governed by an information-bottleneck coefficient $\beta$ that weights the KL regularizer relative to the evidential likelihood~\citep{li2025calibrating_llm}. The optimal $\beta$ does not transfer across model scales or datasets and must be selected from a small grid per (model, dataset) pair. For each dataset, we evaluated 4–5 candidate values spanning roughly two orders of magnitude (Table~\ref{tab:ibedl_beta}), and independently for Base and Sidekick, we picked the candidate that minimized the hold-out ECE on seed 1; the selected entries are bold-faced. The chosen $\beta$ values were then frozen across all 5 repetitions used to produce the IB-EDL rows of Tables~\ref{tab:results_overall} and~\ref{tab:compat}.

\newpage
\begin{table}[H]
\caption{Hold-out ECE (seed 1) of the base and sidekick models across the candidate $\beta$ values evaluated for each dataset. The bold entry in each (dataset, model) column is the value used in the main experiments.}
\centering
\label{tab:ibedl_beta}
\footnotesize
\setlength{\tabcolsep}{10pt}
\begin{tabular}{lccc}
\toprule
Dataset & $\beta$ & Base ECE & Sidekick ECE \\
\midrule
\multirow{4}{*}{ARC-C}
  & $9 \times 10^{-5}$ & \textbf{0.0439} & 0.1118 \\
  & $1.5 \times 10^{-4}$ & 0.0467 & 0.0986 \\
  & $5 \times 10^{-4}$ & 0.1207 & \textbf{0.0538} \\
  & $1.0 \times 10^{-3}$ & 0.1633 & 0.0898 \\
\midrule
\multirow{4}{*}{ARC-E}
  & $2 \times 10^{-6}$ & \textbf{0.0424} & 0.0548 \\
  & $2 \times 10^{-5}$ & 0.0684 & \textbf{0.0508} \\
  & $9 \times 10^{-5}$ & 0.1138 & 0.0693 \\
  & $1.0 \times 10^{-4}$ & 0.1065 & 0.0796 \\
\midrule
\multirow{5}{*}{CSQA}
  & $5 \times 10^{-5}$ & 0.0577 & 0.0638 \\
  & $9 \times 10^{-5}$ & \textbf{0.0501} & \textbf{0.0417} \\
  & $1.5 \times 10^{-4}$ & 0.0686 & 0.0505 \\
  & $5 \times 10^{-4}$ & 0.1224 & 0.0968 \\
  & $1.0 \times 10^{-3}$ & 0.1515 & 0.1283 \\
\midrule
\multirow{5}{*}{OBQA}
  & $6 \times 10^{-5}$ & 0.0812 & \textbf{0.0560} \\
  & $9 \times 10^{-5}$ & \textbf{0.0811} & 0.1150 \\
  & $1.5 \times 10^{-4}$ & 0.0880 & 0.0710 \\
  & $5 \times 10^{-4}$ & 0.1357 & 0.1122 \\
  & $1.0 \times 10^{-3}$ & 0.1606 & 0.1283 \\
\midrule
\multirow{4}{*}{RACE}
  & $1 \times 10^{-6}$ & 0.0425 & 0.0477 \\
  & $2 \times 10^{-6}$ & \textbf{0.0405} & 0.0476 \\
  & $7 \times 10^{-6}$ & 0.0526 & \textbf{0.0461} \\
  & $9 \times 10^{-5}$ & 0.0976 & 0.0791 \\
\midrule
\multirow{4}{*}{SciQ}
  & $5 \times 10^{-7}$ & 0.0300 & \textbf{0.0386} \\
  & $1 \times 10^{-6}$ & \textbf{0.0179} & 0.0493 \\
  & $2 \times 10^{-6}$ & 0.0400 & 0.0476 \\
  & $9 \times 10^{-5}$ & 0.1082 & 0.0924 \\
\bottomrule
\end{tabular}
\end{table}

\end{document}